\documentclass[conference]{IEEEtran}
\IEEEoverridecommandlockouts
\usepackage{cite}
\usepackage{amsmath,amssymb,amsfonts}
\usepackage[ruled,vlined]{algorithm2e}
\usepackage{algorithmic}
\usepackage{pgfplots}
\pgfplotsset{compat=1.14}
\usepackage{textcomp}
\usepackage{multirow}
\usepackage{booktabs}
\usepackage{adjustbox}
\usepackage{soul}
\usepackage{dirtytalk}
\usepackage{mwe,units}
\usepackage{acronym}
\usepackage{ragged2e}
\usepackage{makecell}
\usepackage{efbox,graphicx}
\efboxsetup{linecolor=black,linewidth=0.5pt}

\begin{document}

\title{Brain-inspired self-organization with cellular neuromorphic computing for multimodal unsupervised learning}

\author{
\IEEEauthorblockN{Lyes Khacef, Laurent Rodriguez, Beno\^it Miramond}
\IEEEauthorblockA{\textit{Universit\'e C\^ote d'Azur, CNRS, LEAT, France} \\
firstname.lastname@univ-cotedazur.fr}}
\maketitle


\begin{abstract}
Cortical plasticity is one of the main features that enable our ability to learn and adapt in our environment. Indeed, the cerebral cortex self-organizes itself through structural and synaptic plasticity mechanisms that are very likely at the basis of an extremely interesting characteristic of the human brain development: the multimodal association.
In spite of the diversity of the sensory modalities, like sight, sound and touch, the brain arrives at the same concepts (convergence). Moreover, biological observations show that one modality can activate the internal representation of another modality when both are correlated (divergence).
In this work, we propose the Reentrant Self-Organizing Map (ReSOM), a brain-inspired neural system based on the reentry theory using Self-Organizing Maps and Hebbian-like learning. We propose and compare different computational methods for unsupervised learning and inference, then quantify the gain of the ReSOM in a multimodal classification task. 
The divergence mechanism is used to label one modality based on the other, while the convergence mechanism is used to improve the overall accuracy of the system. We perform our experiments on a constructed written/spoken digits database and a DVS/EMG hand gestures database.
The proposed model is implemented on a cellular neuromorphic architecture that enables distributed computing with local connectivity. We show the gain of the so-called hardware plasticity induced by the ReSOM, where the system's topology is not fixed by the user but learned along the system's experience through self-organization.
\end{abstract}



\section{Introduction} 
\label{sec_intro}
\footnote{Words: 10400; Figures: 11; Tables: 4.}
Intelligence is often defined as the ability to adapt to the environment through learning. "A person possesses intelligence insofar as he has learned, or can learn, to
adjust himself to his environment", S. S. Colvin quoted in \cite{sternberg2000intelligence}. The same definition could be applied to machines and artificial systems in general.
Hence, a stronger relationship with the environment is a key challenge for future intelligent artificial systems that interact in the real-world environment for diverse applications like object detection and recognition, tracking, navigation, etc. The system becomes an "agent" in which the so-called intelligence would emerge from the interaction it has with the environment, as defined in the embodiement hypothesis that is widely adopted in both developmental psychology \cite{smith2005embodied_cognition} and developmental robotics \cite{droniou2015multimodal_perception}. In this work, we tackle the first of the six fundamental principles for the development of embodied intelligence as defined in \cite{smith2005embodied_cognition}: the multimodality.

Indeed, biological systems perceive their environment through diverse sensory channels: vision, audition, touch, smell, proprioception, etc. The fundamental reason lies in the concept of degeneracy in neural structures \cite{edelman1987neural_darwinism}, which is defined by Edelman as the ability of biological elements that are structurally different to perform the same function or yield the same output \cite{edelman2001degeneracy}. In other words, it means that any single function can be carried out by more than one configuration of neural signals, so that the system still functions with the loss of one component. It also means that sensory systems can educate each other, without an external teacher \cite{smith2005embodied_cognition}.
The same principles can be applied for artificial systems, as information about the same phenomenon in the environment can be acquired from various types of sensors: cameras, microphones, accelerometers, etc. Each sensory-information can be considered as a modality. Due to the rich characteristics of natural phenomena, it is rare that a single modality provides a complete representation of the phenomenon of interest \cite{lahat2015multimodal_overview}.

Multimodal data fusion is thus a direct consequence of the well-accepted paradigm that certain natural processes and phenomena are expressed under completely different physical guises \cite{lahat2015multimodal_overview}. Recent works show a growing interest toward multimodal association in several applicative areas such as developmental robotics \cite{droniou2015multimodal_perception}, audio-visual signal processing \cite{shivappa2010audio-visual} \cite{rivet2010speech_separation}, spacial perception \cite{pitti2012gain-field} \cite{fiack2015bio-inspired_vision}, attention-driven selection \cite{braun2019attention-driven} and tracking \cite{zhao2019dynamic_fusion}, memory encoding \cite{tan2019memory_encoding}, emotion recognition \cite{zhang2019deep_fusion}, human-machine interaction \cite{turk2014multimodal_interaction}, remote sensing and earth observation \cite{debes2014hyperspectral_lidar}, medical diagnosis \cite{hoeks2011prostate}, understanding brain functionality \cite{horwitz2002eeg_meg}, etc. Interestingly, the last mentioned application is our starting bloc: how does the brain handle multimodal learning in the natural environment? In fact, it is most likely the emergent result of one of the most impressive abilities of the embodied brain: the cortical plasticity which enables self-organization.

In this work, we propose the Reentrent Self-Organizing Map (ReSOM), a new brain-inspired computational model of self-organization for multimodal unsupervised learning in neuromorphic systems. Section \ref{sec_state-of-art} describes the Reentry framework of Edelman \cite{edelman1982reentry} and the Convergence Divergence Zone framework of Damasio \cite{damasio1989cdz}, two different theories in neuroscience for modeling multimodal association in the brain, and then review some of their recent computational models and applications. Section \ref{sec_model} details the proposed ReSOM multimodal learning and inference algorithms, while section \ref{sec_iterative-grid} presents an extension of the Iterative Grid (IG) \cite{rodriguez2018grid_som} which is applied to ditribute the systems's computation in a cellular neuromorphic architecture for FPGA implementations. Then, section \ref{sec_results} presents the databases, experiments and results with the different case studies. Finally, section \ref{sec_discussion} discusses the results and quantifies the gain of the so-called hardware plasticity through self-organization.


\section{Multimodal learning: state of art} 
\label{sec_state-of-art}

\subsection{Brain-inspired approaches: Reentry and Convergence Divergence Zone (CDZ)}

\begin{figure}[ht]
	\centerline{\includegraphics[width=0.7\linewidth]{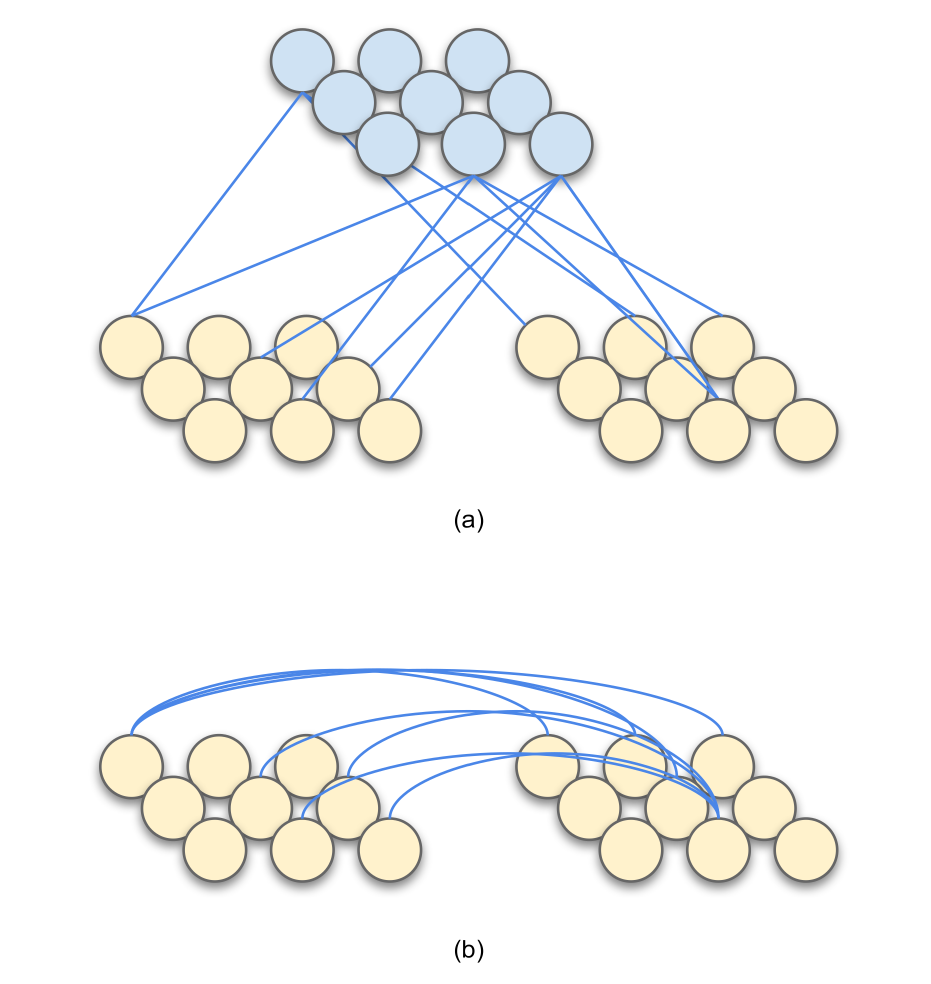}}
	\caption{Schematic representation of (a) CDZ and (b) reentry frameworks. The reentry paradigm states that unimodal neurons connect to each other through direct connections, while the CDZ paradigm implies hierarchical neurons that connect unimodal neurons.}
	\label{fig_reentry-vs-cdz}
\end{figure}

Brain's plasticity, also known as neuroplasticity, is the key to humans capability to learn and modify their behaviour. The plastic changes happen in neural pathways as a result of the multimodal sensori-motor interaction in the environment \cite{escobar-juarez2016soima}. But since most of the stimuli are processed by the brain in more than one sensory modality \cite{meyer2009cdz}, how do the multimodal information \textit{converge} in the brain? Indeed, we can recognize a dog by seeing its picture, hearing its bark or rubbing its fur. These features are different patterns of energy at our sensory organs (eyes, ears and skin) that are represented in specialized regions of the brain. However, we arrive at the same concept of the "dog" regardless of which sensory modality was used \cite{man2015multimodal}. Furthermore, modalities can \textit{diverge} and activate one another when they are correlated. Recent studies have demonstrated cross-modal activation amongst various sensory modalities, like reading words with auditory and olfactory meanings that evokes activity in auditory and olfactory cortices \cite{kiefer2008sound} \cite{gonzalez2006olfactory}, or trying to discriminate the orientation of a tactile grid pattern with eyes closed that induces activity in the visual cortex \cite{sathian2002mind_eye}. Both mechanisms rely on the cerebral cortex as a substrate. But even though recent works have tried to study the human brain’s ability to integrate inputs from multiple modalities \cite{calvert2001crossmodal_processing} \cite{kriegstein2006multisensory_association}, it is not clear how the different cortical areas connect and communicate with each other.

To answer this question, Edelman proposed in 1982 the Reentry \cite{edelman1982reentry} \cite{edelman1993neural_darwinism}: the ongoing bidirectional exchange of signals linking two or more brain areas, one of the most important integrative mechanisms in vertebrate brains \cite{edelman1982reentry}. In a recent review \cite{edelman2013reentry}, Edelman defines reentry as a process which involves a localized population of excitatory neurons that simultaneously stimulates and is stimulated by another population, as shown in Figure \ref{fig_reentry-vs-cdz}. It has been shown that reentrant neuronal circuits self-organize early during the embryonic development of vertebrate brains \cite{singer1990assemblies} \cite{shatz1992thalamus_cortex}, and can give rise to patterns of activity with Winner-Takes-All (WTA) properties \cite{douglas2004neocortex} \cite{rutishauser2009coupled_rnn}. When combined with appropriate mechanisms for synaptic plasticity, the mutual exchange of signals amongst neural networks in distributed cortical areas results in the spatio-temporal integration of patterns of neural network activity. It allows the brain to categorize sensory inputs, remember and manipulate mental constructs, and generate motor commands \cite{edelman2013reentry}. Thus, reentry would be the key to multimodal integration in the brain.

Damasio proposed another answer in 1989 with the Convergence Divergence Zone (CDZ) \cite{damasio1989cdz} \cite{damasio1994cdz}, another biologically plausible framework for multimodal association. In a nutshell, the CDZ theory states that particular cortical areas act as sets of pointers to other areas, with a hierarchical construction: the CDZ merges low level cortical areas with high level amodal constructs, which connects multiple cortical networks to each other and therefore solves the problem of multimodal integration. The CDZ convergence process integrates unimodal information into multimodal areas, while the CDZ divergence process propagates the multimodal information to the unimodal areas, as shown in Figure \ref{fig_reentry-vs-cdz}. For example, when someone talks to us in person, we simultaneously hear the speaker’s voice and see the speaker’s lips move. As the visual movement and the sound co-occur, the CDZ would associate (convergence) the respective neural representations of the two events in early visual and auditory cortices into a higher cortical map. Then, when we only watch a specific lip movement without any sound, the activity pattern induced in the early visual cortices would trigger the CDZ and the CDZ would retro-activate (divergence) in early auditory cortices the representation of the sound that usually accompanied the lip movement \cite{meyer2009cdz}. 

The bidirectionality of the connections is therefore a fundamental characteristic of both reentry and CDZ frameworks, that are likewise in many aspects. Indeed, we find computational models of both paradigms in the literature. We review the most significant ones to our work in Section \ref{sec_statte-of-art_models&apps}.

\subsection{Models and applications}
\label{sec_statte-of-art_models&apps}
In this section, we review the recent works that explore brain-inspired multimodal learning for two main applications: sensori-motor mapping and multi-sensory classification.

\begin{table*}[ht]
\centering
\caption{Models and applications of brains-inspired multimodal learning}
\label{tab_summary}
\begin{center}
\resizebox{0.85\linewidth}{!}{
\begin{tabular}{l l l l l}
\hline
\textbf{Application}                                                   & \textbf{Work}                           & \textbf{Paradigm} & \textbf{Learning}        & \textbf{Computing}                  \\ \hline
\multirow{4}{*}{\makecell{Sensori-motor \\ mapping}} & Lallee et al. \cite{lallee2013mmcm} (2013)             & CDZ      & Unsupervised    & Centralized                \\ 
                                                              & Droniou et al. \cite{droniou2015multimodal_perception} (2015)            & CDZ      & Unsupervised    & Centralized                \\
                                                              & Escobar-Juarez et al. \cite{escobar-juarez2016soima} (2016)     & CDZ      & Unsupervised    & Centralized                \\
                                                              & Zahra et al. \cite{zahra2019sensorimotor} (2019)              & Reentry  & Unsupervised    & Centralized                \\ \hline
\multirow{4}{*}{\makecell{Multi-sensory \\ classification}}     & Parisi et al. \cite{parisi2017multimodal_action} (2017)             & Reentry  & Semi-supervised & Centralized                \\
                                                              & Jayaratne et al. \cite{jayaratne2018multi-sensory_fusion} (2018)          & Reentry  & Semi-supervised & Distributed (data level)   \\
                                                              & Rathi et al. \cite{rathi2018multimodal_stdp} (2018)             & Reentry  & Unsupervised      & Centralized $^{**}$        \\
                                                              & Cholet et al. \cite{cholet2019associative_memory} (2019)             & Reentry $^*$ & Supervised      & Centralized                \\
                                                              & Khacef et al. [this work] (2020) & Reentry  & Unsupervised    & Distributed (system level) \\ \hline
\end{tabular}
}
\end{center}
\begin{flushleft}
    \footnotesize{$^*$ with an extra layer for classification. \\
    $^{**}$ learning is distributed but inference for classification is centralized.}
\end{flushleft}
\end{table*}

\subsubsection{Sensori-motor mapping}
Lallee and Dominey \cite{lallee2013mmcm} proposed the MultiModal Convergence Map (MMCM) that applies the Self-Organizing Map (SOM) \cite{kohonen1990som} to model the CDZ framework. 
The MMCM was applied to encode the sensori-motor experience of a robot based on the language, vision and motor modalities. This "knowledge" was used in return to control the robot behaviour, and increase its performance in the recognition of its hand in different postures.
A quite similar approach is followed Escobar-Juarez et al. \cite{escobar-juarez2016soima} who proposed the Self-Organized Internal Models Architecture (SOIMA) that models the CDZ framework based on internal models \cite{wolpert1998internal_models}. The necessary property of bidirectionality is pointed out by the authors. SOIMA relies on two main learning mechanisms: the first one consists in SOMs that create clusters of unimodal information coming from the environment, and the second one codes the internal models by means of connections between the first maps using Hebbian learning \cite{hebb1949theory} that generates sensory–motor patterns.
A different approach is used by Droniou et al. \cite{droniou2015multimodal_perception} where the authors proposed a CDZ model based on Deep Neural Neteworks (DNNs), which is used in a robotic platform to learn a task from proprioception, vision and audition.
Following the reentry paradigm, Zahra et al. \cite{zahra2019sensorimotor} proposed the Varying Density SOM (VDSOM) for characterizing sensorimotor relations in robotic systems with direct bidirectional connections. 
The proposed method relies on SOMs and associative properties through Oja's learning \cite{oja1982neuron_model} that enables it to autonomously obtain sensori-motor relations without prior knowledge of either the motor (e.g. mechanical structure) or perceptual (e.g. sensor calibration) models.

\subsubsection{Multi-sensory classification}
Parisi et al. \cite{parisi2017multimodal_action} proposed a hierarchical architecture with Growing When Required (GWR) networks \cite{marsland2002gwr} for learning human actions from audiovisual inputs. The neural architecture consists of a self-organizing hierarchy with four layers of GWR for the unsupervised processing of visual action features. The fourth layer of the network implements a semi-supervised algorithm where action–word mappings are developed via the direct bidirectional connections, following the reentry paradigm.
With the same paradigm, Jayaratne et al. \cite{jayaratne2018multi-sensory_fusion} proposed a multisensory neural architecture of multiple layers of Growing SOMs (GSOM) \cite{alahakoon2000gsom} and inter-sensory associative connections representing the co-occurrence probabilities of the modalities. The system's principle is to supplement the information on a single modality with the corresponding information on other modalities for a better classification accuracy.
Using spike coding, Rathi and Roy \cite{rathi2018multimodal_stdp} proposed an STDP-based multimodal unsupervised learning for Spiking Neural Networks (SNNs), where the goal is to learn the cross-modal connections between areas of single modality in SNNs to improve the classification accuracy and make the system robust to noisy inputs. Each modality is represented by a specific SNN trained with its own data following the learning framework proposed in \cite{diehl2015stdp}, and cross-modal connections between the two SNNs are trained along with the unimodal connections. The proposed method was experimented with a written/spoken digits classification task, and the collaborative learning results in an accuracy improvement of $2 \%$. The work of Rathi and Roy \cite{rathi2018multimodal_stdp} is the closest to our work, we threfore confront it in section \ref{sec_som-vs-snn}.
Finally, Cholet et al. \cite{cholet2019associative_memory} proposed a modular architecture for multimodal fusion using Bidirectional Associative Memories (BAMs). First, unimodal data are processed by as many independent Incremental Neural Networks (INNs) \cite{Azcarraga1991incremental_network} as the number of modalities, then multiple BAMs learn pairs of unimodal prototypes. Finally, a INN performs supervised classification.

\subsubsection{Summary}
Overall, the reentry and CDZ frameworks share two key aspects: the multimodal associative learning based on the temporal co-occurrence of the modalities, and the bidirectionality of the associative connections. 
We summarize the most relevant papers to our work in Table \ref{tab_summary}, where we classify each paper with respect to the application, the brain-inspired paradigm, the learning type and the computing nature. 
We notice that sensori-mapping is based on unsupervised learning, which is natural as no label is necessary to map two modalities together. However, classification is based on either supervised or semi-supervised learning, as mapping multi-sensory modalities is not sufficient: we need to know the corresponding class to each activation pattern. We proposed in \cite{khacef2019self-organizing_neurons} a labeling method summarized in Section \ref{sec_som-labeling} based on very few labeled data, so that we do not use any label in the learning process as explained in Section \ref{sec_post-labeled-som}. The same approach is used in \cite{rathi2018multimodal_stdp}, but the authors rely on the complete labeled dataset, as further discussed in section \ref{sec_som-vs-snn}. 
Finally, all previous works rely on the centralized Von Neumann computing paradigm, except \cite{jayaratne2018multi-sensory_fusion} that attempts a partially distributed computing with respect to data, i.e. using the MapReduce computing paradigm to speed up computations. It is based on Apache Spark \cite{gu2013hadoop_spark}, mainly used for cloud computing. Also, STDP learning in \cite{rathi2018multimodal_stdp} is distributed, but the inference for classification requires a central unit, as discussed in Section \ref{sec_som-vs-snn}. We propose a completely distributed computing on the edge with respect to the system, i.e. the neurons computing itself to improve the SOMs scalability for hardware implementation as presented in Section \ref{sec_iterative-grid}.

Consequently, we chose to follow the reentry paradigm where multimodal processing is distributed in all cortical maps without dedicated associative maps for two reasons. 
First, from the brain-inspired computing perspective, more biological evidences tend to confirm the hypothesis of reentry as reviewed by \cite{barth1995rat_cortex}, \cite{allman2009not_just} and \cite{lefort2013somma}. Indeed, biological observations highlight a multimodal processing in the whole cortex including sensory areas \cite{calvert2004multisensory_processing} which contain multimodal neurons that are activated by multimodal stimuli \cite{barth1995rat_cortex} \cite{bizley2008visual_auditory}. Moreover, it has been shown that there are direct connections between sensory cortices \cite{cappe2009multisensory_pathways} \cite{schroeder2005multisensory_contributions}, and neural activities in one sensory area may be influenced by stimuli from other modalities \cite{allman2009not_just} \cite{dehner2004cross-modal}.
Second, from a pragmatical and functional perspective, the reentry paradigm fits better to the cellular architecture detailed in Section \ref{sec_iterative-grid}, and thus increases the scalability and fault tolerance thanks to the completely distributed processing \cite{lefort2013somma}.
Nevertheless, we keep the \textit{convergence} and \textit{divergence} terminology to distinguish between, respectively, the integration of two modalities and the activation of one modality based on the other.


\section{Proposed model: Reentrant Self-Organizing Map (ReSOM)}
\label{sec_model}

\begin{figure}[ht]
	\centerline{\includegraphics[width=1.0\linewidth]{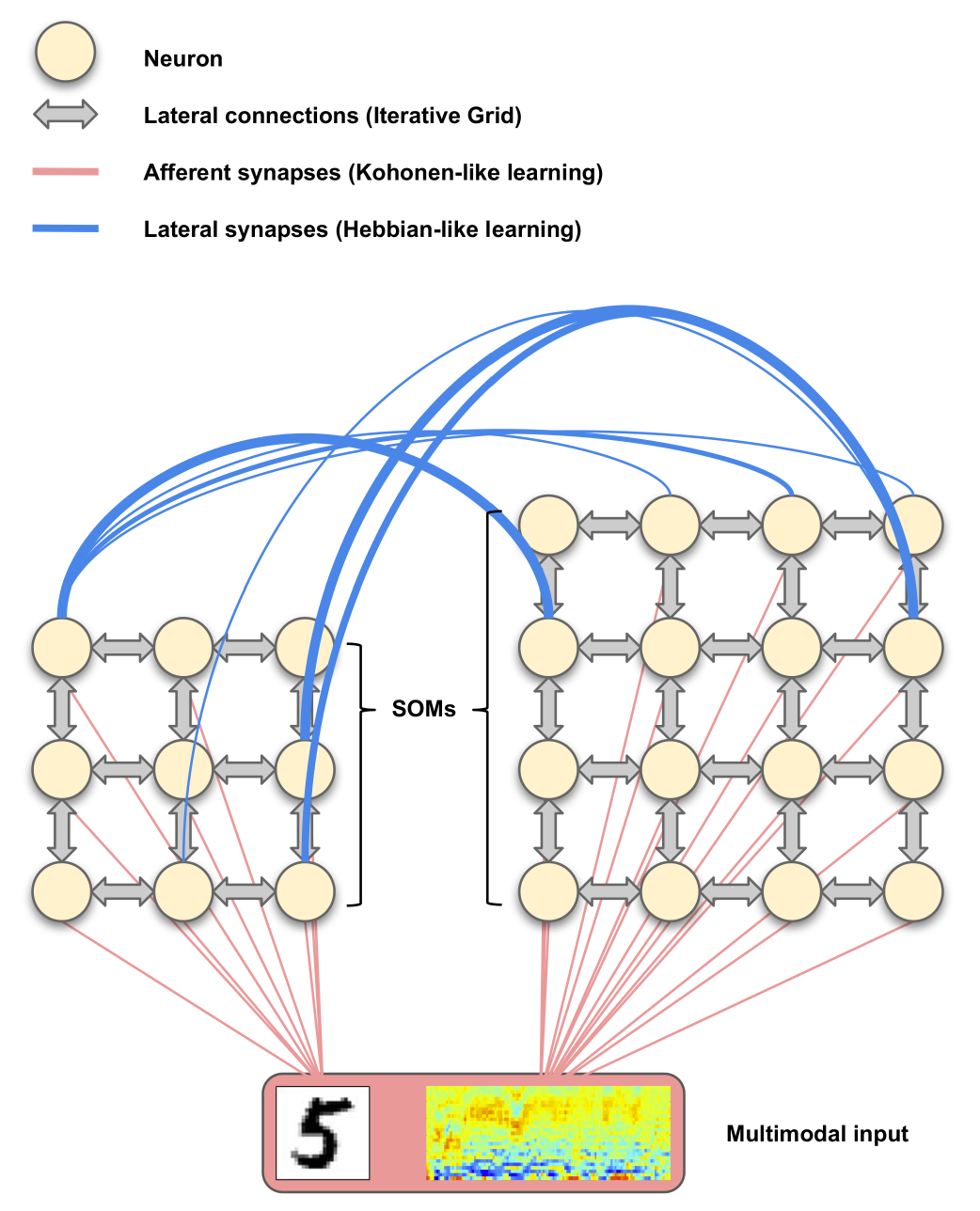}}
	\caption{Schematic representation of the proposed ReSOM for multimodal association. For clarity, the lateral connections of only two neurons from each map are represented.}
	\label{fig_reentry-schematic}
\end{figure}

In this section, we summarise our previous work on SOM post-labeled unsupervised learning \cite{khacef2019self-organizing_neurons}, then propose the Reentrant Self-Organizing Map (ReSOM) shown in Figure \ref{fig_reentry-schematic} for learning multimodal associations, labeling one modality based on the other and converge the two modalities with \textit{cooperation} and \textit{competition} for a better classification accuracy.
We use SOMs and Hebbian-like learning in two times to perform multimodal learning: first, unimodal representations are obtained with SOMs and, second, multimodal representations develop through the association of unimodal maps via bidirectional synapses. 
Indeed, the development of associations between co-occurring stimuli for multimodal binding has been strongly supported by neurophysiological evidence \cite{fiebelkorn2010dual_mechanism}, and follow the reentry paradigm \cite{edelman2013reentry}.

\subsection{Unimodal post-labeled unsupervised learning with Self-Organizing Maps (SOMs)}
\label{sec_post-labeled-som}
With the increasing amount of unlabeled data gathered everyday through Internet of Things (IoT) devices and the difficult task of labeling each sample, DNNs are slowly reaching the limits of supervised learning \cite{droniou2015multimodal_perception} \cite{chum2019beyond_supervised}. Hence, unsupervised learning is becoming one of the most important and challenging topics in Machine Learning (ML) and AI. The Self-Organizing Map (SOM) proposed by Kohonen \cite{kohonen1990som} is one of the most popular Artificial Neural Networks (ANNs) in the unsupervised learning category \cite{kohonen2001som}, inspired from the cortical synaptic plasticity and used in a large range of applications \cite{kohonen1996som_app} going from high-dimensional data analysis to more recent developments such as identification of social media trends \cite{silva2018social_media}, incremental change detection \cite{nallaperuma2018bahavior_changes} and energy consumption minimization on sensor networks \cite{kromes2019lorawan}.
We introduced in \cite{khacef2019self-organizing_neurons} the problem of post-labeled unsupervised learning: no label is available during SOM training then very few labels are available for assigning each neuron the class it represents. The latter is called the labeling phase, which is to distinguish from the fine-tuning process in semi-supervised learning where a labeled subset is used to re-adjust the synaptic weights.

\subsubsection{SOM learning}
The original Kohonen SOM algorithm \cite{kohonen1990som} is described in Algorithm \ref{alg:ksom}. It is to note that $t_f$ is the number of epochs, i.e. the number of times the whole training dataset is presented.
The $\alpha$ hyper-parameter value in Equation \ref{eq_gaussian} is not important for the SOM training, since it does not change the neuron with the maximum activity. It can be set to $1$ in Algorithm \ref{alg:ksom}.
All unimodal trainings were performed over $10$ epochs with the same hyper-parameters as in our previous work \cite{khacef2019self-organizing_neurons}: $\epsilon_i = 1.0$, $\epsilon_f = 0.01$, $\sigma_i = 5.0$ and $\sigma_f = 0.01$.

\begin{algorithm}[h!]
    \caption{SOM unimodal learning}
    \label{alg:ksom}
    \begin{algorithmic}[1]
        \STATE \textbf{Initialize} the network as a two-dimensional array of $k$ neurons, where each neuron $n$ with $m$ inputs is defined by a two-dimensional position $p_n$ and a randomly initialized $m$-dimensional weight vector $w_n$.
        \FOR{$t$ from $0$ to $t_f$}
            \FOR{every input vector $v$}
                \FOR{every neuron $n$ in the network}
                    \STATE \textbf{Compute} the afferent activity $a_n$:
                        \begin{equation}
                        \label{eq_gaussian}
                            a_n = e^{-\frac{||v - w_n||}{\alpha}}
                        \end{equation}
                \ENDFOR
            	\STATE \textbf{Compute} the winner $s$ such that:
                	\begin{equation}
                	\label{eq_max_activity}
                        a_s = \max_{n=0}^{k-1} \left( a_n \right)
                    \end{equation}
                \FOR{every neuron $n$ in the network}
                	\STATE \textbf{Compute} the neighborhood function $h_\sigma(t,n,s)$:
                    \begin{equation}
                    	h_{\sigma}(t,n,s) = e^{-\frac{||p_n - p_s||^2}{2{\sigma}(t)^2}}
                    \end{equation}
                    \STATE \textbf{Update} the weight $w_n$ of the neuron $n$:
                    \begin{equation}
                    	w_n = w_n + \epsilon(t) \times h_{\sigma}(t,n,s) \times (v - w_n)
                    \end{equation}
                \ENDFOR
            \ENDFOR
            \STATE \textbf{Update} the learning rate $\epsilon(t)$:
            \begin{equation}
            	\epsilon(t) = \epsilon_i \left(\frac{\epsilon_f}{\epsilon_i}\right)^{t/t_f}
            \end{equation}
            \STATE \textbf{Update} the width of the neighborhood $\sigma(t)$:
            \begin{equation}
            	\sigma(t) = \sigma_i \left(\frac{\sigma_f}{\sigma_i}\right)^{t/t_f}
            \end{equation}
    	\ENDFOR
    \end{algorithmic}
\end{algorithm}

\subsubsection{SOM labeling}
\label{sec_som-labeling}
The labeling is the step between training and test where we assign each neuron the class it represents in the training dataset. We proposed in \cite{khacef2019self-organizing_neurons} a labeling algorithm based on few labeled samples. We randomly took a labeled subset of the training dataset, and we tried to minimize its size while keeping the best classification accuracy. Our study showed that we only need $1\%$ of randomly taken labeled samples from the training dataset for MNIST \cite{lecun1998mnist} classification.

The labeling algorithm detailed in \cite{khacef2019self-organizing_neurons} can be summarized in five steps.
First, we calculate the neurons activations based on the labeled input samples from the euclidean distance following Equation \ref{eq_gaussian}, where $v$ is the input vector, $w_n$ and $a_n$ are respectively the weights vector and the activity of the neuron $n$. The parameter $\alpha$ is the width of the Gaussian kernel that becomes a hyper-parameter for the method, as further discussed in Section \ref{sec_results}.
Second, the Best Matching Unit (BMU), i.e. the neuron with the maximum activity is elected.
Third, each neuron accumulates its normalized activation (simple division) with respect to the BMU activity in the corresponding class accumulator, and the three steps are repeated for every sample of the labeling subset.
Fourth, each class accumulator is normalized over the number of samples per class.
Fifth and finally, the label of each neuron is chosen according to the class accumulator that has the maximum activity.

\subsection{ReSOM multimodal association: sprouting, Hebbian-like learning and pruning}
Brain's plasticity can be divided into two distinct forms of plasticity: the (1) structural plasticity that changes the neurons connectivity by sprouting (creating) or pruning (deleting) synaptic connections, and (2) the synaptic plasticity that modifies (increasing or decreasing) the existing synapses strength \cite{fauth2016structural_plasticity}. We explore both mechanisms for multimodal association through Hebbian-like learning.

The original Hebbian learning principle \cite{hebb1949theory} proposed by Hebb in 1949 states that \say{when an axon of cell A is near enough to excite a cell B and repeatedly or persistently takes part in firing it, some growth process or metabolic change takes place in one or both cells such that A’s efficiency, as one of the cells firing B, is increased.}
In other words, any two neurons that are repeatedly active at the same time will tend to become "associated" so that activity in one facilitates activity in the other. The learning rule is expressed by Equation \ref{eq_hebb}.

However, Hebb's rule is limited in terms of stability for online learning, as synaptic weights tend to infinity with a positive learning rate. This could be resolved by normalizing each weight over the sum of all the corresponding neuron weights, which guarantees the sum of each neuron weights to be equal to $1$. The effects of weights normalization are explained in \cite{goodhill1994normalization}. However, this solution breaks up with the locality of the synaptic learning rule, and that is not biologically plausible.
In 1982, Oja proposed a Hebbian-like rule \cite{oja1982neuron_model} that adds a "forgetting" parameter, and solves the stability problem with a form of local multiplicative normalization for the neurons weights, as expressed in Equation \ref{eq_oja}. In addition, Oja's learning performs an \textit{on-line} Principal Component Analysis (PCA) of the data in the neural network \cite{fyfe1997pca}, which is a very interesting property in the context of unsupervised learning.

Nevertheless, Hebb's and Oja's rules were both used in recent works with good results, respectively in \cite{escobar-juarez2016soima} and \cite{zahra2019sensorimotor}. Hence, we applied and compared both rules.
The proposed ReSOM multimodal association model is detailed in Algorithm \ref{alg:MultimodalAssociation}, where $\eta$ is a learning rate that we fix to $1$ in our experiments, and $\gamma$ is deduced according to the number or the percentage of synapses to prune, as discussed in Section \ref{sec_results}.
The neurons activities computing in the line $3$ of Algorithm \ref{alg:MultimodalAssociation} are calculated following Equation \ref{eq_gaussian}.

\begin{algorithm}[h!]
    \caption{ReSOM multimodal association learning}
    \label{alg:MultimodalAssociation}
    \begin{algorithmic}[1]
        \STATE \textbf{Learn neurons afferent weights} for $SOM_x$ and $SOM_y$ corresponding to modalities $x$ and $y$ respectively.
        \FOR{every multimodal input vectors $v_x$ and $v_y$}
            \STATE \textbf{Compute} the $SOM_x$ and $SOM_y$ neurons activities.
            \STATE \textbf{Compute} the unimodal BMUs $n_x$ and $n_y$ with activities $a_x$ and $a_y$ respectively.
            \IF{Lateral connection $w_{xy}$ between $n_x$ and $n_y$ does not exist}
                \STATE \textbf{Sprout} (create) the connection $w_{xy} = 0$.
            \ELSE
                \STATE \textbf{Update} lateral connection $w_{xy}$:
                \IF{Hebb's learning}
                    \STATE 
                    \begin{equation}
                    \label{eq_hebb}
                        w_{xy} = w_{xy} + \eta \times a_x \times a_y
                    \end{equation}
                \ELSIF{Oja's learning}
                    \STATE 
                    \begin{equation}
                    \label{eq_oja}
                        w_{xy} = w_{xy} + \eta \times (a_x \times a_y - w_{xy} \times a_y^2)
                    \end{equation}
                \ENDIF
            \ENDIF
        \ENDFOR
        \FOR{every neuron in $SOM_x$}
            \STATE \textbf{Sort} the lateral synapses $w_{xy}$ and deduce the pruning threshold $\gamma$.
            \FOR{every lateral synapse $w_{xy}$}
                \IF{$w_{xy} < \gamma$}
                    \STATE \textbf{Prune} (delete) the connection $w_{xy}$.
                \ENDIF
            \ENDFOR
        \ENDFOR
    \end{algorithmic}
\end{algorithm}

\subsection{ReSOM divergence for labeling}
As explained in \ref{sec_som-labeling}, neurons labeling is based on a labeled subset from the training database. We tried in \cite{khacef2019self-organizing_neurons} to minimize its size, and used the fewest labeled samples while keeping the best accuracy. We will see in Section \ref{sec_results} that depending on the database, we sometimes need a considerable number of labeled samples, up to $10 \%$ of the training set.
In this work, we propose an original method based on the divergence mechanism of the multimodal association: for two modalities $x$ and $y$, since we can activate one modality based on the other, we propose to label the $SOM_y$ neurons from the activity and the labels induced from the $SOM_x$ neurons, which are based on the labeling subset of modality $x$. Therefore, we only need one labeled subset of a single modality which needs the fewest labels to label both modalities, taking profit of the bidirectional aspect of reentry.
A good analogy to biological observations would be the retro-activation of the auditory cortical areas from the visual cortex, if we take the example of written/spoken digits presented in Section \ref{sec_results}. It is similar to how infants respond to sound symbolism by associating shapes with sounds \cite{asano2015sound_symbolism}.
The proposed ReSOM divergence method for labeling is detailed in Algorithm \ref{alg:DivergenceLabeling}.

\begin{algorithm}[h!]
    \caption{ReSOM divergence for labeling}
    \label{alg:DivergenceLabeling}
    \begin{algorithmic}[1]
        \STATE \textbf{Initialize} $class_{act}$ as a two-dimentionnal array of accumulators: the first dimension is the neurons and the second dimension is the classes.
        \FOR{every input vector $v_x$ of the $x$-modality labeling set with label $l$}
            \FOR{every neuron $x$ in the $SOM_x$ map}
                \STATE \textbf{Compute} the afferent activity $a_x$:
                \begin{equation}
                \label{eq_cdz_gaussian_div}
                    a_x = e^{-\frac{||v_x - w_x||}{\beta}}
                \end{equation}
            \ENDFOR
            \FOR{every neuron $y$ in the $SOM_y$ map}
                \STATE \textbf{Compute} the divergent activity $a_y$ from the $SOM_x$:
                \begin{equation}
                \label{eq_cdz_divergence}
                    a_y = \max_{x=0}^{n-1} \left( w_{xy} \times a_x \right)
                \end{equation}
                \STATE \textbf{Add} the normalized activity with respect to the max activity to the corresponding accumulator:
                \begin{equation}
                    class_{act}[y][l] += a_y
                \end{equation}
            \ENDFOR
        \ENDFOR
        \STATE \textbf{Normalize} the accumulators $class_{act}$ with respect to the number of samples per class.
        \FOR{every neuron $y$ in the $SOM_y$ map}
            \STATE Assign the neuron label $neuron_{lab}$:
            \begin{equation}
                neuron_{lab} = argmax(class_{act}[y])
            \end{equation}
        \ENDFOR
    \end{algorithmic}
\end{algorithm}

\subsection{ReSOM convergence for classification}
Once the multimodal learning is done and all neurons from both SOMs are labeled, we need to converge the information of the two modalities to achieve a better representation of the multi-sensory input. Since we use the reentry paradigm, there is no hierarchy in the processing, and the neurons computing is completely distributed based on the Iterative Grid detailed in Section \ref{sec_iterative-grid}. 
We propose an original cellular convergence method in the ReSOM, as detailed in Algorithm \ref{alg:convergence}. We can summarize it in three main steps: 
\begin{itemize}
    \item First, there is an \textit{independent} activity computation (Equation \ref{eq_cdz_gaussian}): each neuron of the two SOMs computes its activity based on the afferent activity from the input.
    \item Second, there is a \textit{cooperation} amongst neurons from different modalities (Equations \ref{eq_cdz_cooperation_max} and \ref{eq_cdz_cooperation_sum}): each neuron updates its afferent activity via a multiplication with the lateral activity from the neurons of the other modality.
    \item Third and finally, there is a global \textit{competition} amongst all neurons (line 19 in Algorithm \ref{alg:convergence}): they all compete to elect a winner, i.e. a global BMU with respect to the two SOMs.
\end{itemize}

\begin{algorithm}[h!]
    \caption{ReSOM convergence for classification}
    \label{alg:convergence}
    \begin{algorithmic}[1]
        \FOR{every multimodal input vectors $v_x$ and $v_y$}
            \STATE \textbf{Do in parallel} every following step inter-changing modality $x$ with modality $y$ and vice-versa:
            \STATE \textbf{Compute} the afferent activities $a_x$ and $a_y$:
            \FOR{every neuron $x$ in the $SOM_x$ map}
                \STATE \textbf{Compute} the afferent activity $a_x$:
                    \begin{equation}
                    \label{eq_cdz_gaussian}
                        a_x = e^{-\frac{||v_x - w_x||}{\beta}}
                    \end{equation}
            \ENDFOR
            \STATE \textbf{Normalize} (min-max) the afferent activities $a_x$ and $a_y$.
            \STATE \textbf{Update} the afferent activities $a_x$ and $a_y$ with the lateral activities based on the associative synapses weights $w_{xy}$:
            \IF{Update with $max_{update}$}
                \FOR{every neuron $x$ in the $SOM_x$ map connected to $n$ neurons from the $SOM_y$ map}
                    \STATE 
                    \begin{equation}
                    \label{eq_cdz_cooperation_max}
                        a_x = a_x \times \max_{x=0}^{n-1} \left( w_{xy} \times a_y \right)
                    \end{equation}
                \ENDFOR
            \ELSIF{Update with $sum_{update}$}
                \STATE 
                \FOR{every neuron $x$ in the $SOM_x$ map connected to $n$ neurons from the $SOM_y$ map}
                    \STATE 
                    \begin{equation}
                    \label{eq_cdz_cooperation_sum}
                        a_x = a_x \times \frac{\sum_{x=0}^{n-1} \left( w_{xy} \times a_y \right)}{n}
                    \end{equation}
                \ENDFOR
            \ENDIF
            \STATE \textbf{Compute} the global BMU with the maximum activity between the $SOM_x$ and the $SOM_y$.
        \ENDFOR
    \end{algorithmic}
\end{algorithm}

We explore different variants of the proposed convergence method regarding two aspects. 
First, both afferent and lateral activities can be taken as raw values or normalized values. We used min-max normalization that is therefore done with respect to the BMU and the Worst Matching Unit (WMU) activities. 
Second, the afferent activities update could be done for all neurons or only the two BMUs. In the second case, the global BMU cannot be another neuron but one of the two local BMUs, and if there is a normalization then it is only done for lateral activities (otherwise, the BMUs activities would be $1$, and the lateral map activity would be the only relevant one). The results of our comparative study are presented and discussed in Section \ref{sec_results}.


\section{Cellular neuromorphic architecture} 
\label{sec_iterative-grid}
The centralized neural models that run on classical computers suffer from the Von-Neumann bottleneck due to the overload of communications between computing memory components, leading to a an over-consumption of time and energy. One attempt to overcome this limitation is to distribute the computing amongst neurons as done in \cite{diehl2015stdp}, but it implies an all-to-all connectivity to calculate the global information, e.g. the BMU. Therefore, this solution does not completely solve the initial problem of scalability.

An alternative approach to solve the scalability problem can be derived from the Cellular Automata (CA) which was originally proposed by John von Neumann \cite{von_neumann1967automata} then formally defined by Stephen Wolfram \cite{wolfram1984cellular_automata}. The CA paradigm relies on locally connected cells with local computing rules which define the new state of a cell depending on its own state and the states of its neighbors. All cells can then compute in parallel as no global information is needed. Therefore, the model is massively parallel and is an ideal candidate for hardware implementations \cite{halbach2004cellular_fpga}.
A recent FPGA implementation to simulate CA in real time has been proposed in \cite{kyparissas2019fpga_automata}, where authors show a speed-up of $51 \times$ compared to a high-end CPU (Intel Core i7-7700HQ) and a comparable performance with recent GPUs with a gain of $10 \times$ in power consumption. With a low development cost, low cost of migration to future devices and a good performance, FPGAs are suited to the design of cellular processors \cite{walsh2012fpga_simd}. Cellular architectures for ANNs were common in early neuromorphic implementations and have recently seen a resurgence \cite{schuman2017neuromorphic_survey}. 
Such implementation is also refered as near-memory computing where one embeds dedicated coprocessors in close proximity to the memory unit, thus getting closer to the Parallel and Distributed Processing (PDP) paradigm \cite{blazewicz2000parallel_distributed} formalized in the theory of ANNs.

An FPGA distributed implementation model for SOMs was proposed in \cite{sousa2017embedded_som}, where the local computation and the information exchange among neighboring neurons enable a global self-organization of the entire network.
Similarly, we proposed in \cite{rodriguez2018grid_som} a cellular formulation of the related neural models which would be able to tackle the full connectivity limitation by iterating the propagation of the information in the network. This particular cellular implementation, named the Iterative Grid (IG), reaches the same behavior as the centralized models but drastically reduces their computing complexity when deployed on hardware. 
Indeed, we have shown in \cite{rodriguez2018grid_som} that the time complexity of the IG is $O(\sqrt{n})$ with respect to the number of neurons $n$ in a squared map, while the time complexity of a centralized implementation is $O(n)$. In addition, the connectivity complexity of the IG is $O(n)$ with respect to the number of of neurons $n$, while the connectivity complexity of a distributed implementation with all-to-all connectivity \cite{diehl2015stdp} is $O(n^2)$.
The principles of the IG are summarized in this section followed by a new SOM implementation over the IG substrata which takes in account the needs of the multimodal association learning and inference.

\subsection{The Iterative Grid (IG) substrata}
Let's consider a 2-dimensional grid shaped Network-on-Chip (NoC). This means that each node (neuron) of the network is physically connected (only) to its four closest neighbors.
At each clock edge, each node reads the data provided by its neighbors and relays it to its own neighbors on the next one. The data is propagated (or broadcasted) in a certain amount of time to all the nodes. The maximum amount of time $T_p$ which is needed to cover all the NoC (worst case reference) depends on its size: for a $N \times M$ grid, $T_p = N + M - 2$. After $T_p$ clock edges, new data can be sent. A set of $T_p$ iterations can be seen as a \textit{wave of propagation}.

For the SOM afferent weights learning, the data to be propagated is the maximum activity for the BMU election, plus its distance with respect to every neuron in the map. The maximum activity is transmitted through the wave of propagation, and the distance to the BMU is computed in the same wave thanks to this finding: \say{When a data is iteratively propagated through a grid network, the propagation time is equivalent to the Manhattan distance between the source and each receiver.} \cite{rodriguez2018grid_som}.

\subsection{Iterative Grid for SOM model}
\label{sec_ig-som}

\begin{figure}[ht]
	\centerline{\includegraphics[width=1.0\linewidth]{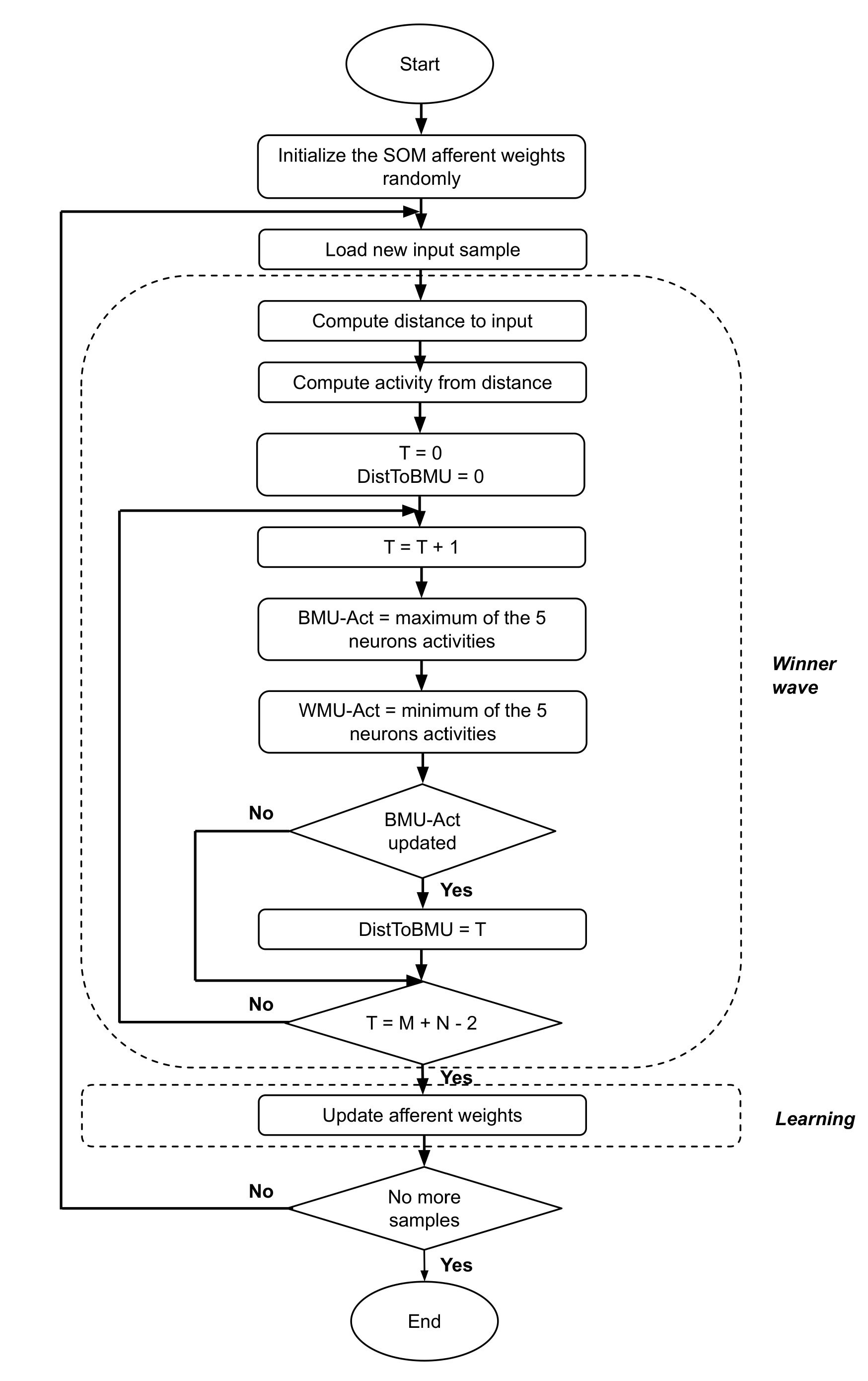}}
	\caption{BMU and WMU distributed computing flowchart for each neuron. This flowchart describes the SOM learning, but the winner wave is applied the same way for all steps of the multimodal learning while the learning part can be replaced by Hebbian-like learning or inference.}
	\label{fig_ig-flowchart}
\end{figure}

The SOM implementation on the IG proposed in \cite{rodriguez2018grid_som} has to be adapted to fit the needs of the multimodal association: (1) we add the Worst Matching Unit (WMU) activity needed for the activities min-max normalization in the convergence step, and (2) we use the Gaussian kernel in Equation \ref{eq_gaussian} to transform the euclidean distances into activities. Therefore, the BMU is the neuron with the maximum activity and the WMU the neuron with the minimum one.
The BMU/WMU search wave called the "winner wave" is described as a flowchart in Figure \ref{fig_ig-flowchart}.
When the BMU/WMU are elected, the next step is the learning wave.
From the winner propagation wave, every useful data is present in each neuron to compute the learning equation. No propagation wave is necessary at this step.

\subsection{Hardware support for the iterative grid}
\label{sec_hw-ig}

\begin{figure}[ht]
	\centerline{\includegraphics[width=1.0\linewidth]{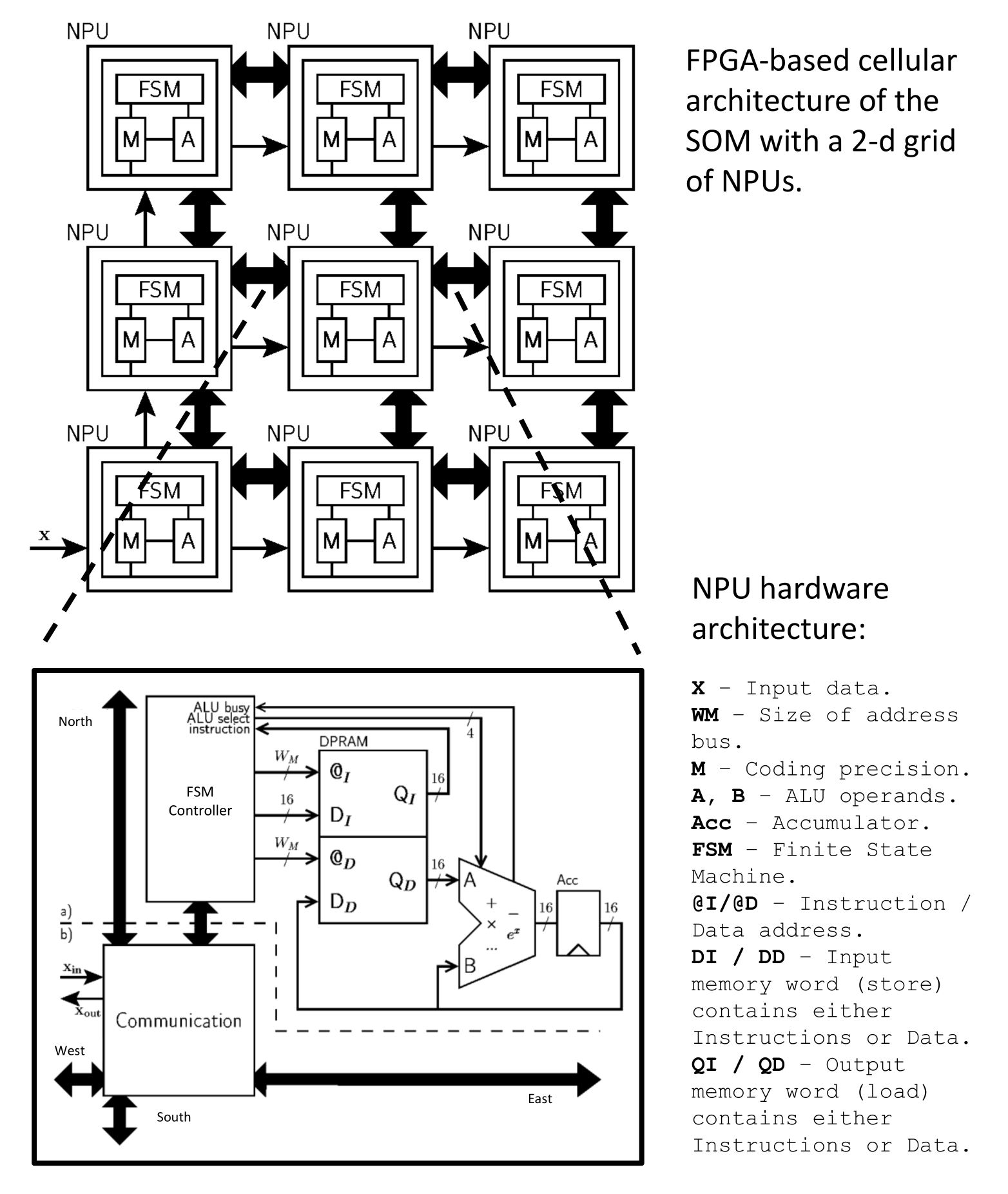}}
	\caption{Neural Processing Units (NPUs) grid on FPGA \cite{fiack2015npu}.}
	\label{fig_ig-npu}
\end{figure}

The multi-FPGA implementation of the IG is a work in progress based on our previously implemented Neural Processing Unit (NPU) \cite{rodriguez2013hardware_plasticity} \cite{fiack2015npu}. As shown in Figure \ref{fig_ig-npu}, the NPU is made of two main parts: the computation core and the communication engine. The computation core is a lightweight Harvard-like accumulator-based micro-processor where a central dual-port RAM memory stores the instructions and the data, both separately accessible from its two ports. A Finite State Machine (FSM) controls the two independent ports of the memory and the Arithmetic and Logic Unit (ALU), which implements the needed operations to perform the equations presented in section \ref{sec_ig-som}. The aim of the communication engine is to bring the input stimuli vector and the neighbors activities to the computation core at each iteration. The values of the input vector flow across the NPUs through their $x_{in}$ and $x_{out}$ ports which are connected as a broadcast tree. The output activity ports of each NPU are connected to the four cardinal neighbors through a dedicated hard-wired channel.

Implemented on an Altera Stratix V GXEA7 FPGA, the resources (LUT, Registers, DSP and memory blocks) consumption is indeed scalable as it increases linearly in function of the size of the NPU network \cite{rodriguez2013hardware_plasticity} \cite{fiack2015npu}. We are currently working on configuring the new model in the NPU and implementing it on a more recent and adapted FPGA device, particularly for the communication part between multiple FPGA boards that will be based on SCALP \cite{vannel2018scalp}.

The cellular approach for implementing SOM models proposed by Sousa et al. \cite{sousa2017embedded_som} is an FPGA implementation that shares the same approach as the IG with distributed cellular computing and local connectivity. However, the IG has two main advantages over the cellular model in \cite{sousa2017embedded_som}:
\begin{itemize}
    \item Waves complexity: The ”smallest of 5” and ”neighborhood” waves in \cite{sousa2017embedded_som} have been coupled into one wave called the ”winner wave”, as the iterative grid is based on time to distance transformation to find the Manhattan distance between the BMU and each neuron. We have therefore a gain of about $2 \times$ in the time complexity of the SOM training.
    
    \item Sequential vs. combinatory architecture: The processes of calculating the neuron distances to the input vector, searching for the BMU and updating the weight vectors are performed in a single clock cycle. This assumption goes against the iterative computing paradigm in the SOM grid to propagate the neurons information. Hence, the hardware implementation in \cite{sousa2017embedded_som} is almost fully combinatory. It explains why the maximum operating frequency is low and decreases when increasing the number of neurons, thus being not scalable in terms of both hardware resources and latency.
\end{itemize}

\subsection{Hardware support for multimodal association}
For the multimodal association learning in Algorithm \ref{alg:MultimodalAssociation}, the local BMU in each of the two SOMs needs both the activity and the position of the local BMU of the other SOM to perform the Hebbian-like learning in the corresponding lateral synapse.
This communication problem has not been experimented in this work. However, this suppose a simple communication mechanism between the two maps that would be implemented in two FPGAs where only the BMUs of each map send a message to each other in a bidirectional way. The message could go through the routers of the IG thanks to an XY-protocol to reach an inter-map communication port in order to avoid the multiplication of communication wires.

For the divergence and convergence methods in Algorithm \ref{alg:DivergenceLabeling} and Algorithm \ref{alg:convergence} respectively, the local BMU in each of the two SOMs needs the activity of all the connected neurons from the other SOM after pruning, i.e. around 20 connections per neuron. Because the number of remaining synapses is statistically bounded to $20 \%$, the number of communication remains low in front of the number of neurons. Here again, we did not experiment on this communication mechanism but the same communication support could be used. Each BMU can send a request that contains a list of connected neurons. This request can be transmitted to the other map through the IG routers to an inter-map communication channel. Once on the other map, the message could be broadcasted to each neuron using again the routers of the IG. Only the requested neurons send back their activity coupled to their position in the BMU request. This simple mechanism supposes a low amount of communication thanks to the pruning that has been done previously. 
This inter-map communication can be possible if the IG routers support XY or equivalent routing techniques and broadcast in addition to the one of the propagation wave.


\section{Experiments and results} \label{sec_results}
In this section, we present the databases and the results from our experiments with each modality alone, then with the multimodal association convergence and divergence, and we finally compare our model to three different approaches. 
All the results presented in this section have been averaged over a minimum of $10$ runs, with shuffled datasets and randomly initialized neurons afferent weights.

\subsection{Databases}
The most important hypothesis that we want to confirm through this work is that the multimodal association of two modalities leads to a better accuracy than the best of the two modalities alone. For this purpose, we worked on two databases that we present in this section.

\subsubsection{Written/spoken digits database}
\label{sec_wr-sp-digits}

\begin{figure*}[ht]
    \centerline{\efbox{\includegraphics[width=1.0\linewidth]{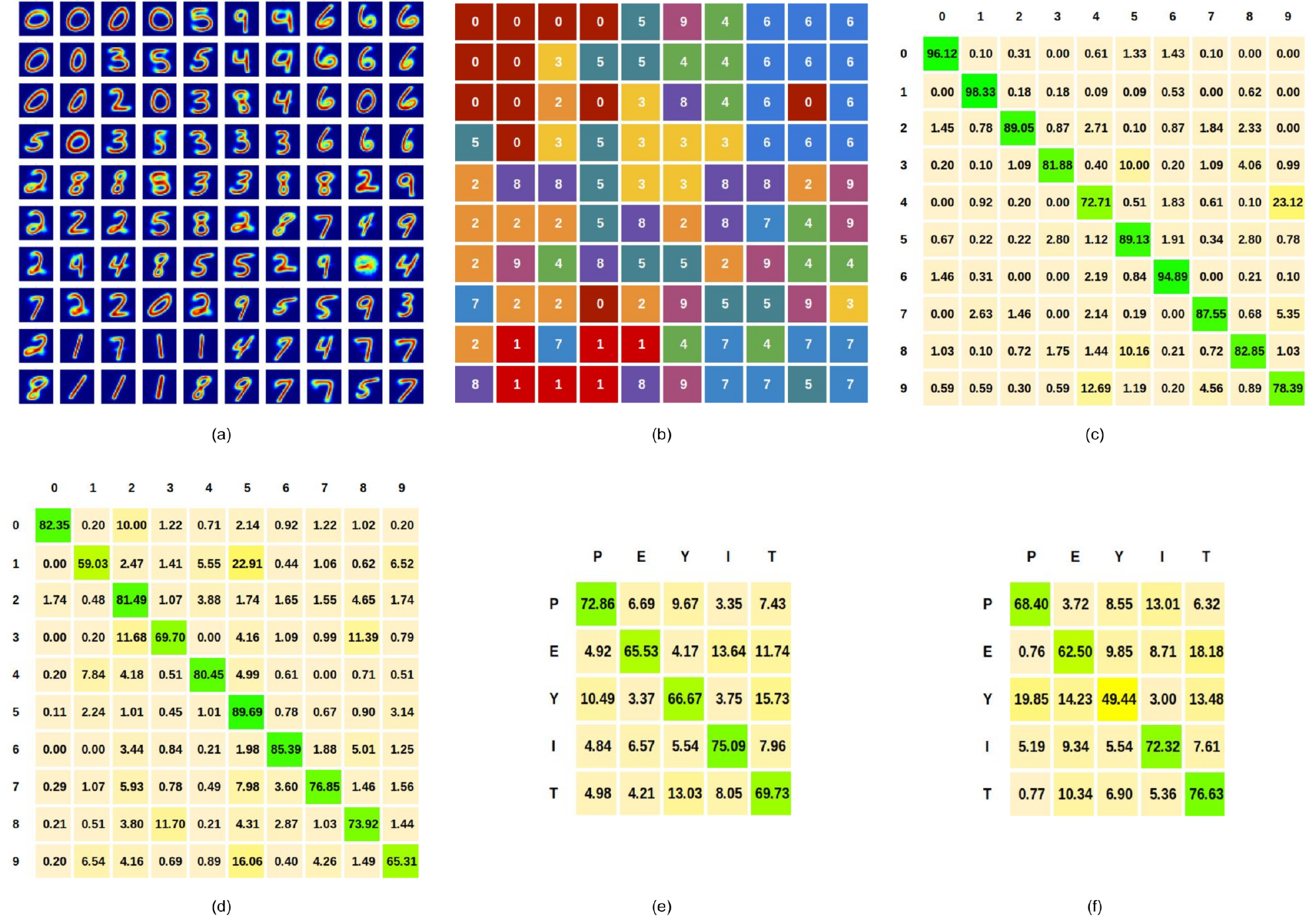}}}
	\caption{MNIST learning with SOM: (a) neurons afferent weights; (b) neurons labels; (c) confusion matrix; we can visually assess the good labeling from (a) and (b), while (c) shows that some classes like $4$ and $9$ are easier to confuse than others, and that's due to their proximity in the 784-dimensional space; (d) S-MNIST divergence confusion matrix; (e) DVS confusion matrix; (f) EMG divergence confusion matrix; the interesting characteristic is that the confusion between the same classes is not the same for the different modalities, and that's why they can complement each other.}
	\label{fig_som-confusion}
\end{figure*}

The MNIST database \cite{lecun1998mnist} is a database of $70000$ handwritten digits ($60000$ for training and $10000$ for test) proposed in 1998. Even if the database is quite old, it is still commonly used as a reference for training, testing and comparing various ML systems for image classification. 
In \cite{khacef2019self-organizing_neurons}, we applied Kohonen-based SOMs for MNIST classification with post-labeled unsupervised learning, and achieved state-of-art performance with the same number of neurons ($100$) and only $1\%$ of labeled samples for the neurons labeling. However, the obtained accuracy of $87.36 \%$ is not comparable to supervised DNNs, and only two approaches have been used in the literature to bridge the gap: either use a huge number of neurons ($6400$ neurons in \cite{diehl2015stdp}) with exponential increase in size for linear increase in accuracy \cite{rathi2018multimodal_stdp} which is not scalable for complex databases, or use unsupervised feature extraction followed by a supervised classifier (Support Vector Machine in \cite{kheradpisheh2018stdp_cnn}) which relies on the complete labeled dataset. 
We propose the multimodal association as a way to bridge the gap while keeping a small number of neurons and an unsupervised learning method from end to end. For this purpose, we use the classical MNIST as a visual modality that we associate to an auditory modality: Spoken-MNIST (S-MNIST).

We extracted S-MNIST from Google Speech Commands (GSC) \cite{warden2018speech_commands}, an audio dataset of spoken words that was proposed in 2018 to train and evaluate keyword spotting systems. It was therefore captured in real-world environments though phone or laptop microphones. The dataset consists of $105829$ utterances of $35$ words, amongst which $38908$ utterances ($34801$ for training and $4107$ for test) of the $10$ digits from $0$ to $9$. We constructed S-MNIST associating written and spoken digits of the same class, respecting the initial partitioning in \cite{lecun1998mnist} and \cite{warden2018speech_commands} for the training and test databases. Since we have less samples in S-MNIST than in MNIST, we duplicated some random spoken digits to match the number of written digits and have a multimodal-MNIST database of $70000$ samples. The whole pre-processed dataset is available in \cite{khacef2019multimodal_digits}.

\subsubsection{DVS/EMG hand gestures database}
To validate our results, we experimented our model on a second database that was originally recorded with multiple sensors: the DVS/EMG hand gestures database \cite{ceolini2019hand_gestures}.
Indeed, the discrimination of human gestures using wearable solutions is extremely important as a supporting technique for assisted living, healthcare of the elderly and neuro-rehabilitation.
For this purpose, we proposed in \cite{ceolini2019dvs_emg_fusion} and and \cite{ceolini2020dvs_emg_neuromorphic} a framework that allows the integration of multi-sensory data to perform sensor fusion based on supervised learning. The framework was applied for the hand gestures recognition task with five hand gestures: Pinky (P), Elle (E), Yo (Y), Index (I) and Thumb (T).

The dataset consists of $6750$ samples ($5400$ for training and $1350$ for test) of muscle activities via EletroMyoGraphy (EMG) signals recorded by a Myo armband (Thalmic Labs Inc) from the forearm, and video recordings from a Dynamic Vision Sensor (DVS) using the computational resources of a mobile phone.
The DVS is an event-based camera inspired by the mammalian retina \cite{lichtsteiner2006128}, such that each pixel responds asynchronously to changes in brightness with the generation of events. Only the active pixels transfer information and the static background is directly removed on hardware at the front-end. The asynchronous nature of the DVS makes the sensor low power, low latency and low-bandwidth, as the amount of data transmitted is very small. It is therefore a promising solution for mobile applications \cite{ceolini2019live_demo_fusion} as well as neuromorphic chips, where energy efficiency is one of the most important characteristics.

\subsection{SOM unimodal classification}

\begin{table*}[h!]
\centering
\caption{Classification accuracies and convergence/divergence gains.}
\label{tab_cdz-gain}
\begin{center}
\resizebox{0.7\linewidth}{!}{
\begin{tabular}{l l c c c c}
\hline
\multicolumn{2}{c}{\multirow{2}{*}{\textbf{Database}}}    & \multicolumn{2}{c}{\textbf{Digits}} & \multicolumn{2}{c}{\textbf{Hand gestures}} \\ \cline{3-6}
\multicolumn{2}{c}{}                             & \textbf{MNIST}           & \textbf{S-MNIST}   & \textbf{DVS}                  & \textbf{EMG}         \\ \hline
\multirow{4}{*}{SOMs} & Dimensions      & 784 & 507 & 972 & 192 \\
                               & Neurons         & 100 & 256 & 256 & 256 \\
                               & Labeled data (\%) & 1               & 10        & 10                   & 10          \\
                               & Accuracy (\%) $_\alpha$     & 87.04 $_{1.0}$           & 75.14 $_{0.1}$     & 70.06 $_{2.0}$               & 66.89 $_{1.0}$       \\ \hline
\multirow{3}{*}{ReSOM Divergence}    & Labeled data (\%) & 1               & 0         & 10                   & 0           \\ 
                               & Gain (\%)         & /               & +0.76      & /                    & -1.33       \\
                               & Accuracy (\%)     & /               & 75.90     & /                    & 65.56       \\ \hline
\multirow{2}{*}{ReSOM Convergence}   & Gain (\%)         & \textbf{+8.03}   & +19.17     & \textbf{+5.67}        & +10.17       \\
                               & Accuracy (\%)     & \multicolumn{2}{c}{95.07}  & \multicolumn{2}{c}{75.73}         \\ \hline
\end{tabular}
}
\end{center}
\end{table*}

\subsubsection{Written digits}
\label{sec_results_mnist}
MNIST classification with a SOM was already performed in \cite{khacef2019self-organizing_neurons}, achieving around $87\%$ of classification accuracy using $1\%$ of labeled images from the training dataset for the neurons labeling. 
The only difference is the computation of the $\alpha$ in Equation \ref{eq_gaussian} for the labeling process. We proposed in \cite{khacef2019self-organizing_neurons} a centralized method for computing an approximated value of $\alpha$, but we consider it as a simple hyper-parameter for this work. We therefore calculate the best value off-line with a grid search since we do not want to include any centralized computation, and because we can find a closer value to the optimum, as summarized in Table \ref{tab_cdz-gain}. The same procedure with the same hyper-parameters defined above is applied for each of the remaining unimodal classifications. 
Finally, we obtain $87.04\% \pm 0.64$ of accuracy. Figure \ref{fig_som-confusion} shows the neurons weights that represent the learned digits prototypes with the corresponding labels, and the confusion matrix that highlights the most frequent misclassifications between the digits whose representations are close:  $23.12 \%$ of the digits $4$ are classified as $9$ and $12.69 \%$ of the digits $9$ are classified as a $4$. We find the same mistakes with a lower percentage between the digits $3$, $5$ and $8$, because of their proximity in the 784-dimensional vector space. That's what we aim to compensate when we add the auditory modality.

\subsubsection{Spoken digits}
The most commonly used acoustic feature in speech recognition is the Mel Frequency Cepstral Coefficients (MFCC) \cite{luque2018mfcc} \cite{darabkh2018mfcc} \cite{pan2018mfcc}. MFCC was first proposed in \cite{mermelstein1976mfcc}, which has since become the standard algorithm for representing speech features. It is a representation of the short-term power spectrum of a speech signal, based on a linear cosine transform of a log power spectrum on a nonlinear Mel scale of frequency.
We first extracted the MFCC features from the S-MNIST data, using the hyper-parameters from \cite{pan2018mfcc}: framing window size = $50\unit{ms}$ and frame shift size = $25\unit{ms}$. Since the S-MNIST samples are approximately $1 s$ long, we end up with $39$ dimensions.
However, it's not clear how many coefficients one has to take. Thus, we compared three methods: \cite{chapaneri2012mfcc} proposed to use $13$ weighted MFCC coefficients, \cite{sainath2015mfcc} proposed to use $40$ log-mel filterbank features, and \cite{pan2018mfcc} proposed to use $12$ MFCC coefficients with an additional energy coefficient, making it $13$ coefficients in total. The classification accuracy is respectively $61.79\% \pm 1.19$, $50.33\% \pm 0.59$ and $75.14\% \pm 0.57$.
We therefore chose to work with a $39 \times 13$ dimensional features that are standardized (each feature is transformed by subtracting the mean value and dividing by the standard deviation of the training dataset, also called Z-score normalization) then min-max normalized (each feature is re-scaled to $0 - 1$ based on the minimum and maximum values of the training dataset).
The confusion matrix in Figure \ref{fig_som-confusion} shows that the confusion between the digits $4$ and $9$ is almost zero, which strengthens our hypothesis that the auditory modality can complement the visual modality for a better overall accuracy.

\subsubsection{DVS hand gestures}
In order to use the DVS events for gesture classification with conventional algorithms, we need to turn the stream of events into frames, as previously done in \cite{ceolini2019dvs_emg_fusion}. These frames are generated by accumulating the events occurring in a fixed time window of $200\unit{ms}$, so that they can be synchronized with the EMG signal.
For each pixel, we count the number of events within the time windows regardless of their polarity, then we transform the event count frame into gray scale by min-max normalization.
The event frames obtained from the DVS camera have a resolution of $128 \times 128$ pixels. 
Since the region with the hand gestures does not fill the full frame, we extract a $60 \times 60$ pixels patch that allows us to significantly decrease the amount of computation needed during learning and inference.

Even though unimodal classification accuracies are not the first goal, we need to reach a \textit{satisfactory} performance. Since the dataset is small and the DVS frames are of high complexity with a lot of noise from the data acquisition, we either have to significantly increase the number of neurons for the SOM or to add an additional feature-extraction. We decided to use the second method with a Convolutional Neural Network (CNN)-based feature extraction to keep a reasonable number of neurons. 
One of our future improvements is to use unsupervised learning for feature extraction based on recent works in \cite{kheradpisheh2018stdp_cnn} and \cite{falez2019feature_learning}, but it is out of the scope of this work. 
Hence, we use supervised feature extraction based on the LeNet-5 topology \cite{lecun1998lenet} with one difference: the last convolution layer has only $12$ filters instead of $120$. In this way, we end up with extracted features of $972$ dimensions, that we standardize and normalize before using as input for the SOM. We obtain an accuracy of $70.06 \% \pm 1.15$.

\subsubsection{EMG hand gestures}
For the EMG signal, we selected two time domain features that are commonly used in the literature \cite{phinyomark2018feature}: the Mean Absolute Value (MAV) and the Root Mean Square (RMS) which are calculated over the same window of length $20\unit{ms}$, as detailed in \cite{ceolini2019dvs_emg_fusion}.
With the same strategy as for DVS frames, we extract CNN-based features of $192$ dimensions. The SOM reaches a classification accuracy of $66.89 \% \pm 0.84$.

\subsection{ReSOM multimodal classification}

\begin{figure}[h!]
    \centerline{\efbox{\includegraphics[width=1.0\linewidth]{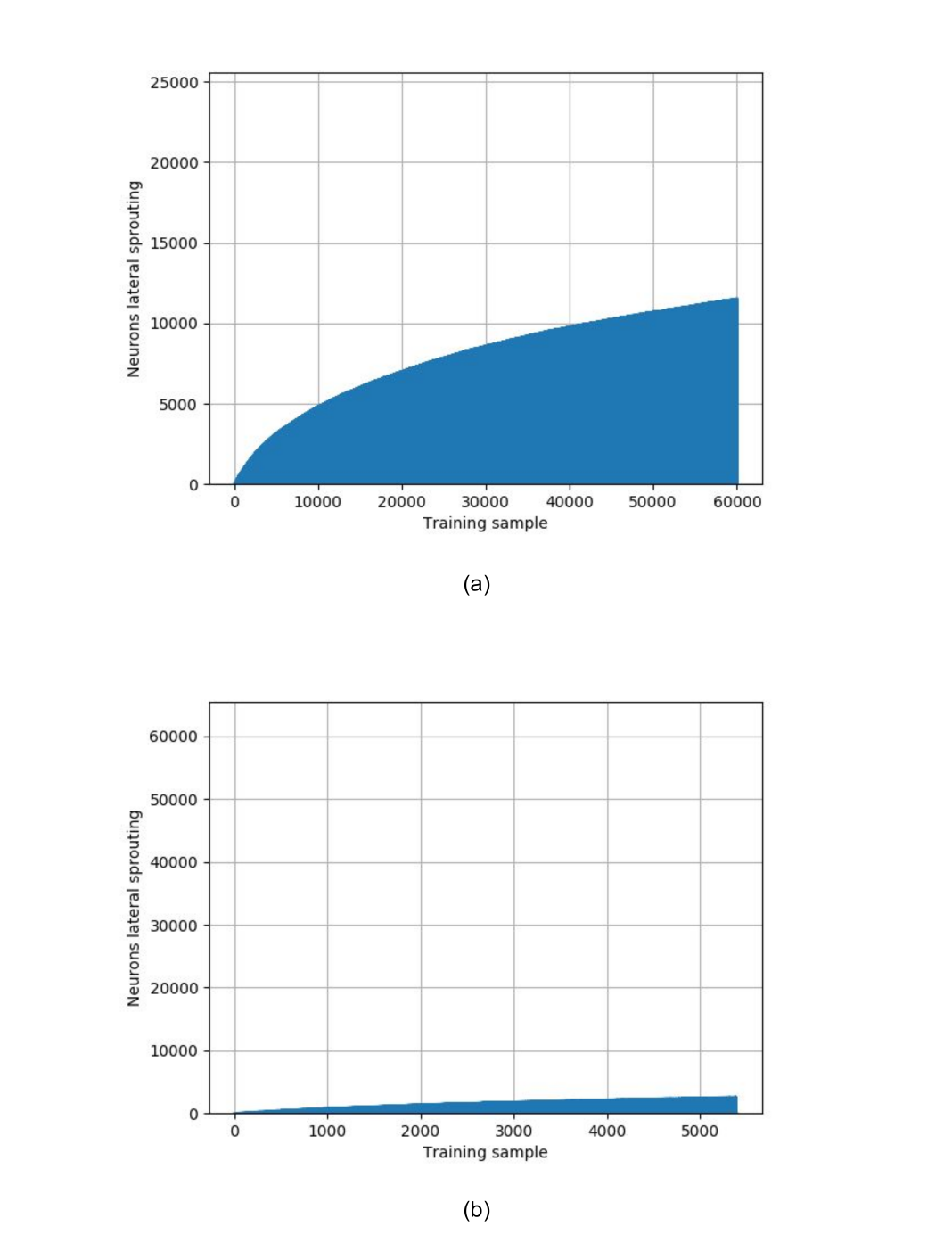}}}
	\caption{SOMs lateral sprouting in the multimodal association process: (a) Written/Spoken digits maps; (b) DVS/EMG hand gestures maps. We notice that less than half of the possible lateral connections are created at the end of the Hebbian-like learning, because only meaningful connections between correlated neurons are created. For (b), the even smaller number of connections is also related to the small size of the training dataset.}
	\label{fig_cdz-sprout}
\end{figure}

\begin{figure}[h!]
    \centerline{\efbox{\includegraphics[width=1.0\linewidth]{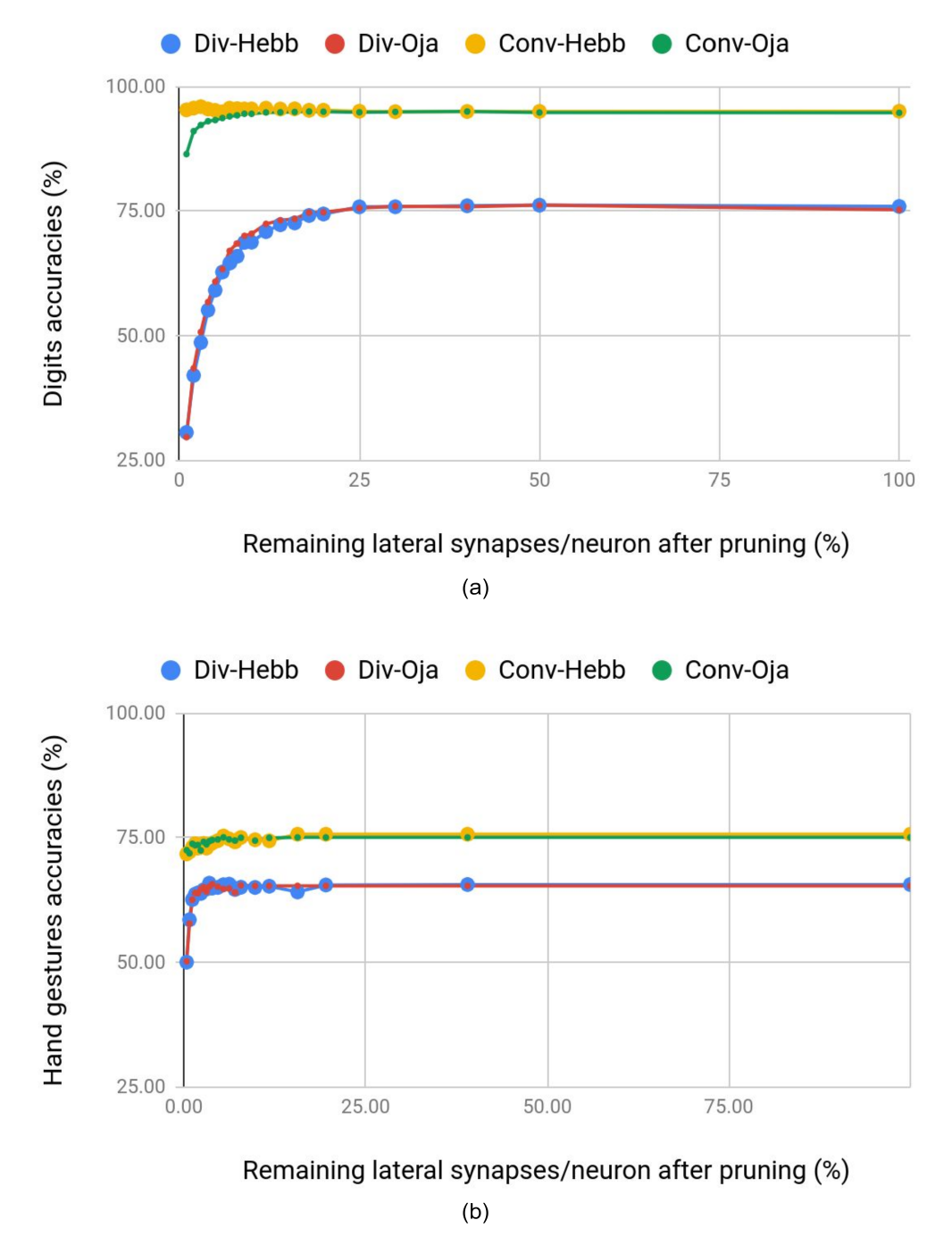}}}
	\caption{Divergence and convergence classification accuracies VS. the remaining percentage of lateral synapses after pruning: (a) Written/Spoken digits maps; (b) DVS/EMG hand gestures maps. We see that we need more connections per neuron for the divergence process, because the pruning is done by the neurons of one of the two maps, and a small number of connections results in some disconnected neurons in the other map.}
	\label{fig_cdz-prune}
\end{figure}

\begin{figure*}[h!]
    \centerline{\efbox{\includegraphics[width=0.8\linewidth]{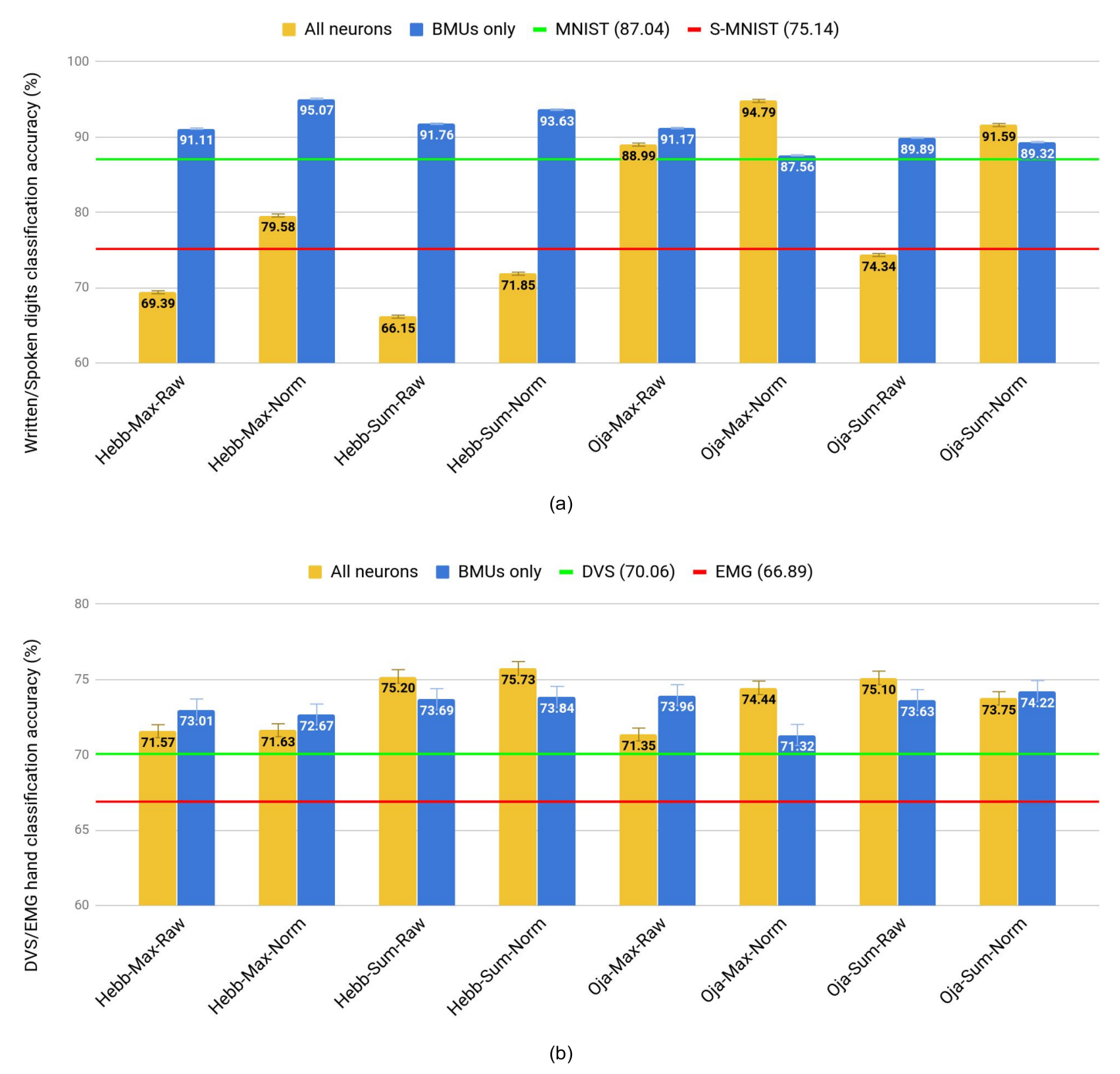}}}
	\caption{Multimodal convergence classification: (a) Written/Spoken digits; (b) DVS/EMG hand gestures. The red and green lines are respectively the lowest and highest unimodal accuracies. Hence, there is an overall gain whenever the convergence accuracy is above the green line.}
	\label{fig_cdz-convergence}
\end{figure*}

\begin{table*}[h!]
\centering
\caption{Multimodal classification accuracies}
\label{tab_multimodal_accuracy}
\begin{center}
\resizebox{0.75\linewidth}{!}{
\begin{tabular}{c c c c c c c}
\hline
\multirow{3}{*}{\textbf{Learning}} & \multicolumn{6}{c}{\textbf{ReSOM convergence method and accuracy (\%) $_\beta$}}                                                                               \\ \cline{2-7} 
                          & \multirow{2}{*}{\makecell{\textbf{Update} \\ \textbf{algorithm}}} & \multirow{2}{*}{\makecell{\textbf{Neurons} \\ \textbf{activities}}} & \multicolumn{2}{c}{\textbf{Digits}}     & \multicolumn{2}{c}{\textbf{Hand gestures}} \\ \cline{4-7} 
                          &                       &                             & \textbf{All neurons}    & \textbf{BMUs only}      & \textbf{All neurons}        & \textbf{BMUs only}     \\ \hline
\multirow{4}{*}{Hebb}     & \multirow{2}{*}{Max}  & Raw                         & 69.39 $_{1}$         & 91.11 $_{1}$          & 71.57 $_{5}$              & 73.01 $_{5}$         \\ 
                          &                       & Norm                        & 79.58 $_{20}$          & \textbf{95.07} $_{10}$ & 71.63 $_{3}$              & 72.67 $_{20}$         \\ \cline{2-7} 
                          & \multirow{2}{*}{Sum}  & Raw                         & 66.15 $_{1}$          & 91.76 $_{10}$          & \textbf{75.20} $_{4}$     & 73.69 $_{4}$         \\ 
                          &                       & Norm                        & 71.85 $_{1}$          & 93.63 $_{20}$          & \textbf{75.73} $_{4}$     & 73.84 $_{20}$         \\ \hline
\multirow{4}{*}{Oja}      & \multirow{2}{*}{Max}  & Raw                         & 88.99 $_{4}$          & 91.17 $_{1}$          & 71.35 $_{3}$              & 73.96 $_{10}$         \\
                          &                       & Norm                        & \textbf{94.79} $_{4}$ & 87.56 $_{3}$          & 74.44 $_{30}$              & 71.32 $_{10}$         \\ \cline{2-7} 
                          & \multirow{2}{*}{Sum}  & Raw                         & 74.34 $_{2}$          & 89.89 $_{3}$          & 75.10 $_{4}$              & 73.63 $_{10}$         \\
                          &                       & Norm                        & 91.59 $_{15}$          & 89.32 $_{30}$          & 73.75 $_{4}$              & 74.22  $_{30}$         \\ \hline
\end{tabular}
}
\end{center}
\end{table*}

After inter-SOM sprouting (Figure \ref{fig_cdz-sprout}), training and pruning (Figure \ref{fig_cdz-prune}), we move to the inference for two different tasks: (1) labeling one SOM based on the activity of the other (divergence), and (2) classifying multimodal data with cooperation and competition between the two SOMs (convergence).

\subsubsection{ReSOM divergence results}
Table \ref{tab_cdz-gain} shows unimodal classification accuracies using the divergence mechanism for labeling, with $75.9\% \pm 0.2$ for S-MNIST classification and $65.56\% \pm 0.25$ for EMG classification. As shown in Figure \ref{fig_cdz-prune}, we reach this performance using respectively $20\%$ and $25\%$ of the potential synapses for digits and hand gestures. Since the pruning is performed by the neurons of the \textit{source} SOMs, i.e. the MNIST-SOM and DVS-SOM, pruning too much synapses causes some neurons of the S-MNIST-SOM and EMG-SOM to be completely disconnected from the source map, and therefore do not get any activity for the labeling process. Hence, the labeling is incorrect, with the disconnected neurons stuck with the default label $0$.
In comparison to the classical labeling process with $10\%$ of labeled samples, we have a loss of only $-1.33\%$ for EMG and  even a small gain of $0.76\%$ for S-MNIST even though we only use $1\%$ of labeled digits images.
The choice of which modality to use to label the other is made according to two criteria: the source map must (1) achieve the best unimodal accuracy so that we maximize the separability of the transmitted activity to the other map, and it must (2) require the least number of labeled data for its own labeling so that we minimize the number of samples to label during data acquisition.
Overall, the divergence mechanism for labeling leads to approximately the same accuracy than the classical labeling. Therefore, we perform the unimodal classification of S-MNIST and EMG with no labels from end to end.

\subsubsection{ReSOM convergence results}

\begin{figure*}[ht]
    \centerline{\efbox{\includegraphics[width=0.75\linewidth]{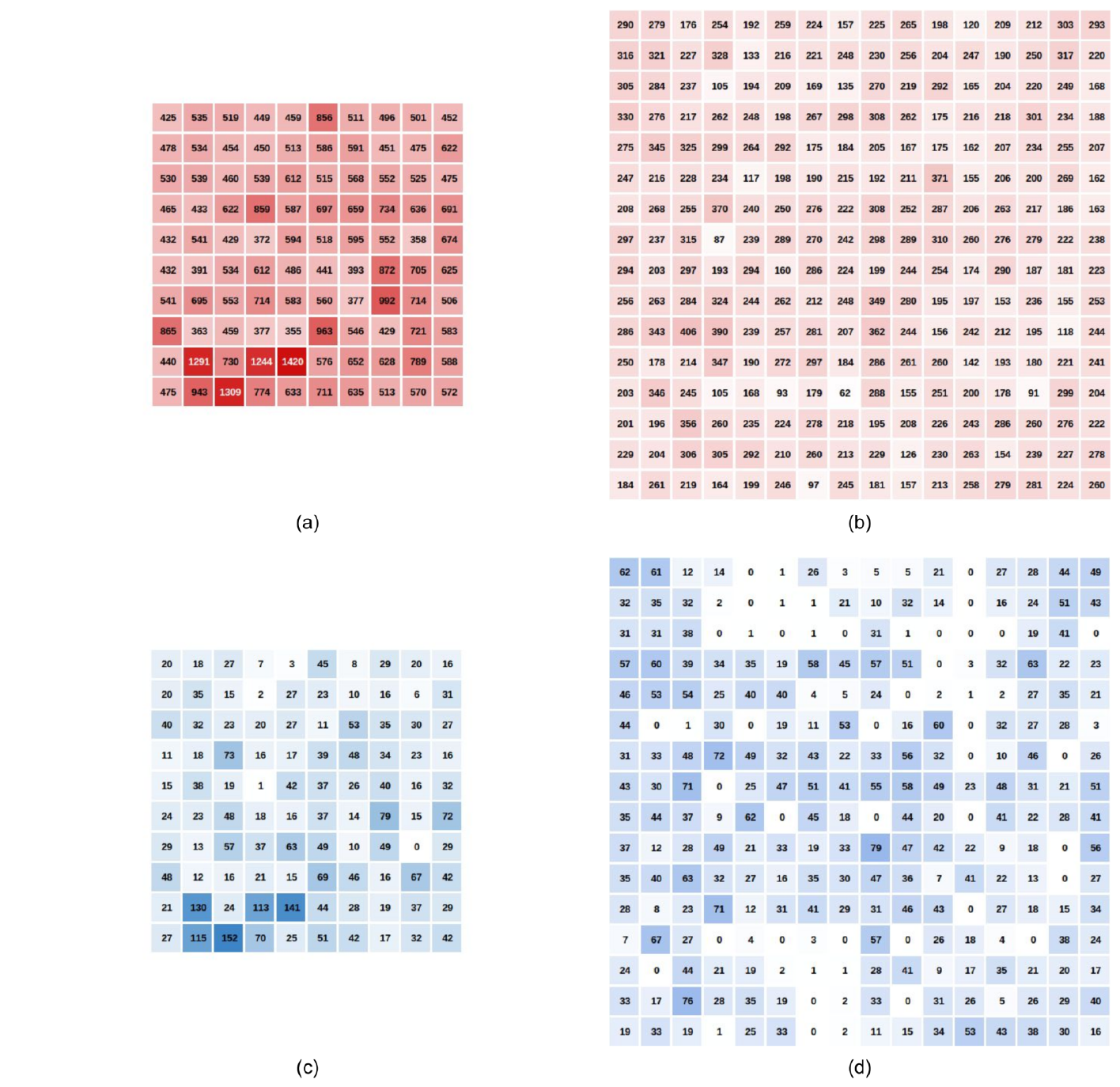}}}
	\caption{Written/Spoken digits neurons BMU counters during multimodal learning and inference using $Hebb-Max_{Norm}^{BMU}$ method: (a) MNIST SOM in learning; (b) S-MNIST SOM neurons during learning; (c) MNIST SOM neurons during inference; (d) S-MNIST SOM neurons during inference.}
	\label{fig_cdz-mnist-cnt}
\end{figure*}

\begin{figure*}[ht]
    \centerline{\efbox{\includegraphics[width=0.75\linewidth]{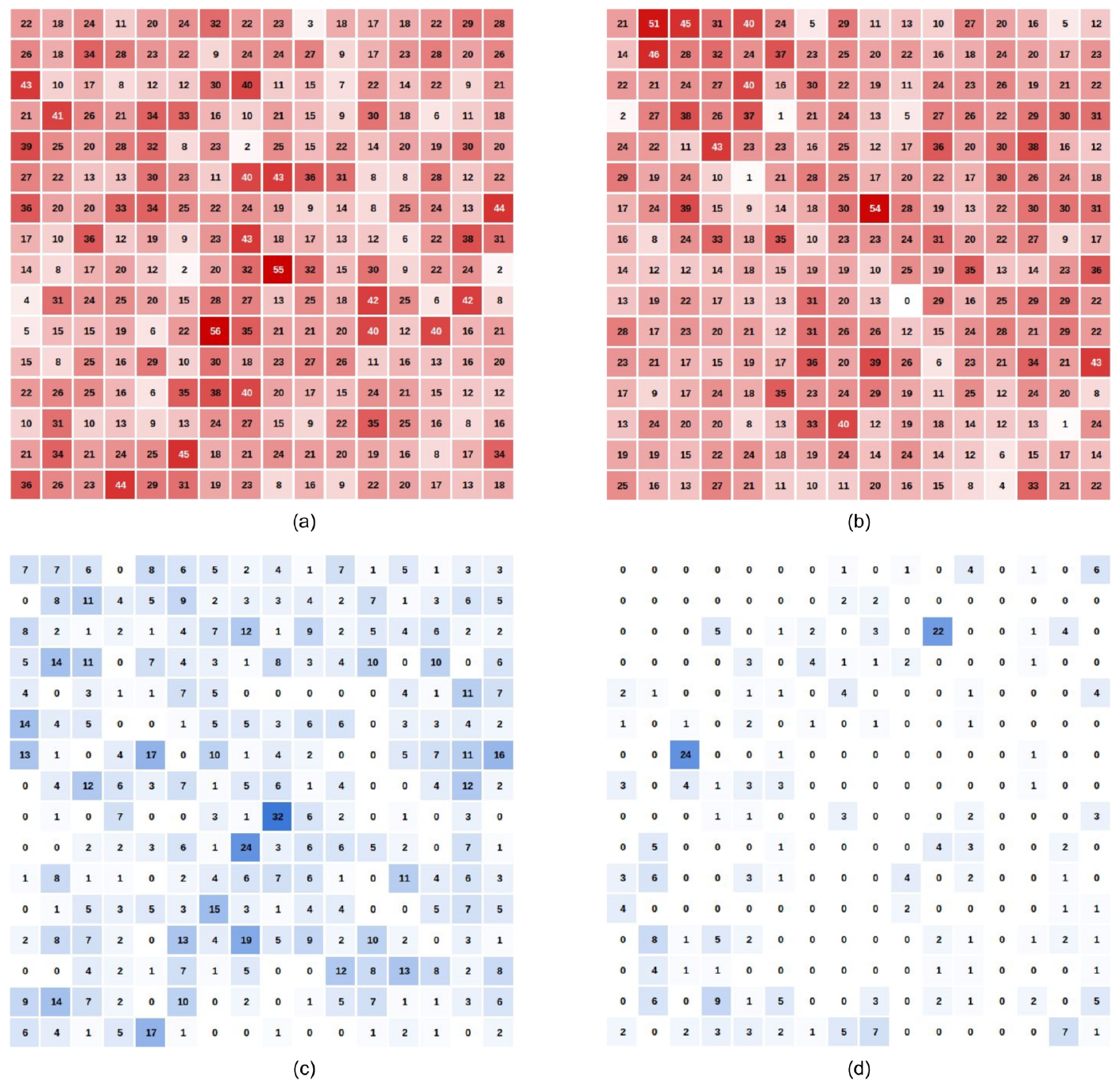}}}
	\caption{DVS/EMG hand gestures neurons BMU counters during multimodal learning and inference using $Hebb-Sum_{Norm}^{All}$ method: (a) DVS SOM in learning; (b) EMG SOM neurons during learning; (c) DVS SOM neurons during inference; (d) EMG SOM neurons during inference.}
	\label{fig_cdz-hand-cnt}
\end{figure*}

\begin{figure*}[ht]
    \centerline{\efbox{\includegraphics[width=1.0\linewidth]{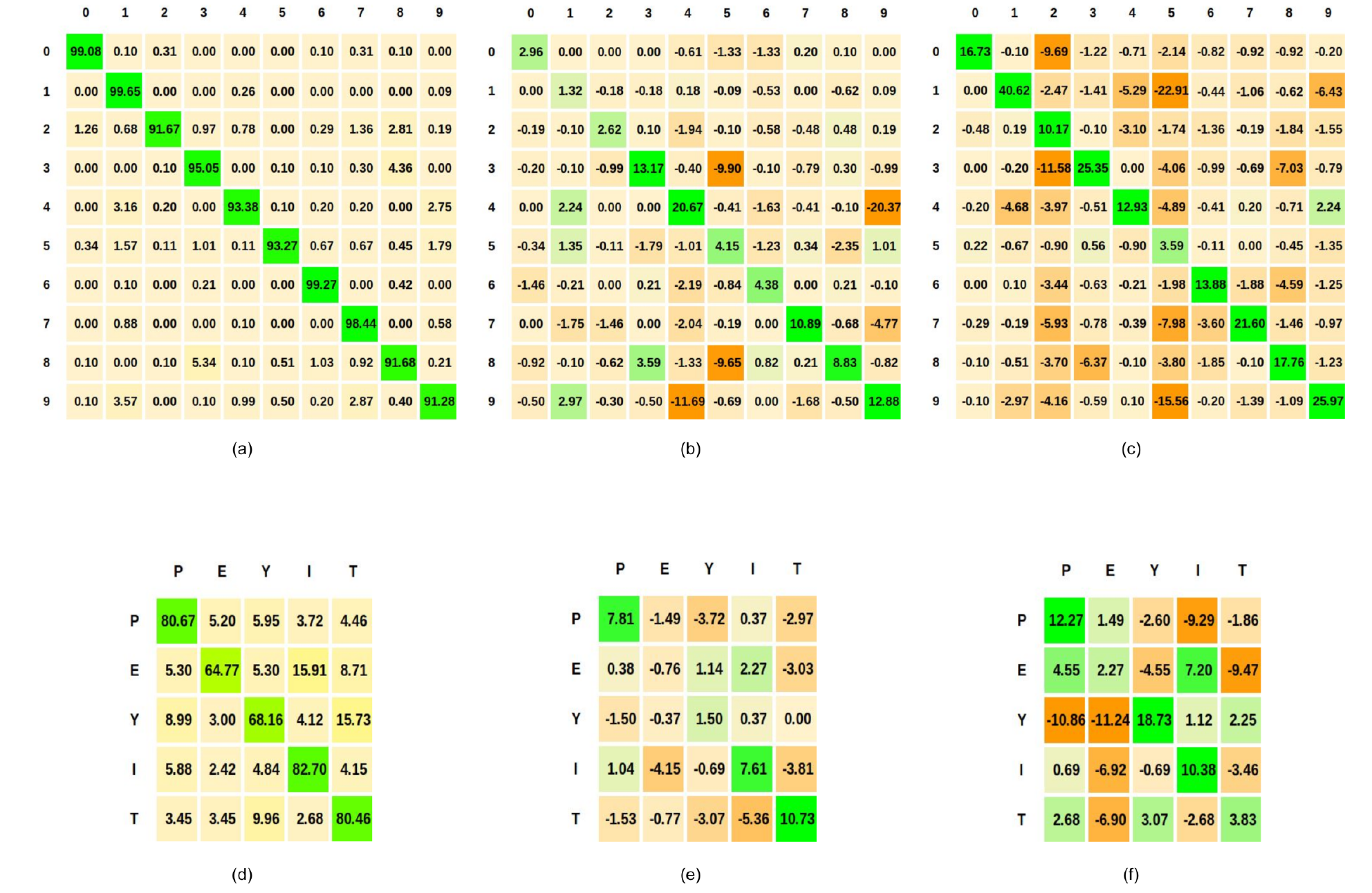}}}
	\caption{Written/Spoken digits confusion matrices using $Hebb-Max_{Norm}^{BMU}$ method: (a) convergence; (b) convergence gain with respect to MNIST; (c) convergence gain with respect to S-MNIST; DVS/EMG hand gestures confusion matrices using $Hebb-Sum_{Norm}^{All}$ method: (d) convergence; (e) convergence gain with respect to DVS; (f) convergence gain with respect to EMG.}
	\label{fig_cdz-confusion}
\end{figure*}

We proposed eight variants of the convergence algorithm for each the two learning methods. For the discussion, we denote them as follow: $Learning-Update_{Normalization}^{Neurons}$ such that $Learning$ can be $Hebb$ or $Oja$, $Update$ can be $Max$ or $Sum$, $Normalization$ can be $Raw$ (the activites are taken as initially computed by the SOM) or $Norm$ (all activities are normalized with a min-max normalization thanks to the WMU and BMU activities of each SOM), and finally $Neurons$ can be $BMU$ (only the two BMUs update each other and all other neurons activities are reset to zero) or $All$ (all neurons update their activities and therefore the global BMU can be different from the two local BMUs).
It is important to note that since we constructed the written/spoken digits dataset, we maximized the cases where the two local BMUs have different labels such as one of them is correct. This choice was made in order to better asses the accuracies of the methods based on BMUs update only, as both cases when the two BMUs are correct or incorrect at the same time lead to the same global results regardless of the update method.
The convergence accuracies for each of the eight method applied on the two databases are summarized in Table \ref{tab_multimodal_accuracy} and Figure \ref{fig_cdz-convergence}.

For the digits, we first notice that the Hebb's learning with all neurons update leads to very poor performance, worse than the unimodal classification accuracies. 
To explain this behavior, we have to look at the neurons BMU counters during learning in Figure \ref{fig_cdz-mnist-cnt}. We notice that some neurons, labeled as $1$ in Figure \ref{fig_som-confusion}, are winners much more often that other neurons. Hence, their respective lateral synapses weights increase disproportionately compared to other synapses, and lead those neurons to be winners most of the time after the update, as their activity is higher than other neurons very often during convergence. 
This behavior is due to two factors: first, the neurons that are active most of the time are those that are the fewest to represent a class. Indeed, there are less neurons labeled $1$ compared to other classes, because the digit $1$ have less \textit{sub-classes}. In other words, the digit $1$ has less variants and therefore can be represented by less prototype neurons. Consequently, those neurons are active more often because the number of samples for each class is approximately equal. 
Second, the Hebb's learning is unbounded, leading the lateral synapses weights to increase indefinitely. Thus, this problem occurs less when we use Oja's rule, as shown in Figure \ref{fig_cdz-convergence}. We notice that Oja's learning leads to more homogenous results, and normalization often leads to a better accuracy. The best method using Hebb's learning is $Hebb-Max_{Norm}^{BMU}$ with $95.07 \% \pm 0.08$, while the best method using Oja's learning is $Oja-Max_{Norm}^{All}$ with $94.79 \% \pm 0.11$.

For the hand gestures, all convergence methods lead to a gain in accuracy even though the best gain is smaller than for digits, as summarized in Table \ref{tab_cdz-gain}. 
It can be explained by the absence of neurons that would be BMUs much more often than other neurons, as shown in Figure \ref{fig_cdz-hand-cnt}. The best method using Hebb's learning is $Hebb-Sum_{Norm}^{All}$ with $75.73 \% \pm 0.91$, while the best method using Oja's learning is $Oja-Sum_{Raw}^{All}$ with $75.10 \% \pm 0.9$. In contrast with the digits database, here the most accurate methods are based on the $Sum$ update. 
Thus, each neuron takes in account the activities of all the neurons that it is connected to. A plausible reason is the fact that the digits database was constructed whereas the hand gestures database was initially recorded with multimodal sensors, which gives it a more natural correlation between the two modalities.

Overall, the best methods for both digits and hand gestures databases are based on Hebb's learning, even though the difference with the best methods based on Oja's learning is very small, and Oja's rule has the interesting property of bounding the synaptic weights. For hardware implementation, the synaptic weights of the Hebb's learning can be normalized after a certain threshold without affecting the model's behavior, since the strongest synapse stays the same when we divide all the synapses by the same value. However, the problem is more complex in the context of on-line learning as discussed in Section \ref{sec_discussion}.
Quantitatively, we have a gain of $+8.03 \%$ and $+5.67 \%$ for the digits and the hand gestures databases respectively, compared to the best unimodal accuracies. The proposed convergence mechanism leads to the election of a global BMU between the two unimodal SOMs: it is one of the local BMUs for the $Hebb-Max_{Norm}^{BMU}$ method used for digits, whereas it can be a completely different neuron for the $Hebb-Sum_{Norm}^{All}$ used for hand gestures. 
In the first case, since the convergence process can only elect one of the two local BMUs, we can compute the \textit{absolute accuracy} in the cases where the two BMUs are different with one of them being correct. We find that the correct choice between the two local BMUs is made in about $87\%$ of the cases.
However, in both cases, the convergence leads to the election a global BMU that is indeed spread in the two maps, as shown in Figures \ref{fig_cdz-mnist-cnt} and \ref{fig_cdz-hand-cnt}. Nevertheless, the neurons of the hand gestures SOMs are less active in the inference process, because we only have $1350$ samples in the test database.

The best accuracy for both methods is reached using a sub-part of the lateral synapses, as we prune a big percentage of the potential synapses as shown in Figure \ref{fig_cdz-prune}. We say \textit{potential} synapses, because the pruning is performed with respect to a percentage (or number) of synapses for each neuron, and the neuron does not have the information of other neurons due to the cellular architecture. Thus, the percentage is calculated with respect to the maximum number of potential lateral synapses, that is equal to the number of neurons in the other SOM, and not the actual number of synapses. In fact, at the end of the Hebbian-like learning, each neuron is only connected to the neurons where there is at least one co-occurrence of BMUs, as shown in Figure \ref{fig_cdz-sprout}. Especially for the hand gestures database, the sprouting leads to a small total number of lateral synapses even before pruning, because of the small number of samples in the training dataset.
Finally, we need at most $10\%$ of the total lateral synapses to achieve the best performance in convergence as shown in Figure \ref{fig_cdz-prune}. However, if we want to maintain the unimodal classification with the divergence method for labeling, then we have to keep $20 \%$ and $25 \%$ of the potential synapses for digits and hand gestures, respectively.

One interesting aspect of the multimodal fusion is the explainability of the better accuracy results. To do so, we plot the confusion matrices with the best convergence methods for the digits and hand gestures datasets in Figure \ref{fig_cdz-confusion}. The gain matrices mean an improvement over the unimodal performance when they have positive values in the diagonal and negative values elsewhere. 
If we look at the gain matrix of the convergence method compared to the image modality, we notice two main characteristics: first, all the values in the diagonal are positive, meaning that there is a total accuracy improvement for all the classes. 
Second and more interestingly, the biggest absolute values outside the diagonal lie where there is the biggest confusion for the images, i.e. between the digits $4$ and $9$, and between the digits $3$, $5$ and $8$, as previously pointed out in Section \ref{sec_results_mnist}. 
It confirms our initial hypothesis, which means that the auditory modality brings a complementary information that leads to a greater separability for the classes which have the most confusion in the visual modality. 
Indeed, the similarity between written $4$ and $9$ is compensated by the dissimilarity of spoken $4$ and $9$. The same phenomenon can be observed for the auditory modality, where there is an important gain for the digit $9$ that is often misclassified as $1$ or $5$ in the speech SOM, due to the similarity of their sounds. 
Similar remarks are applicable for the hand gestures database with more confusion in some cases, which leads to a smaller gain.

Our results confirm that multimodal association is interesting because the strengths and weaknesses of each modality can be complementary. Indeed, Rathi and Roy \cite{rathi2018multimodal_stdp} state that if the non-idealities in the unimodal datasets are independent, then the probability of misclassification is the product of the misclassification probability of each modality. Since the product of two probabilities is always lower than each probability, then each modality helps to overcome and compensate for the weaknesses of the other modality. Furthermore, multimodal association improves the robustness of the overall system to noise \cite{rathi2018multimodal_stdp}, and in the extreme case of losing one modality, the system could rely on the other one which links back to the concept of degeneracy in neural structures \cite{edelman1987neural_darwinism}.

\subsection{Comparative study}
First, we compare our results with STDP approaches to assess the classification accuracy with a comparable number of neurons. Next, we confront our results with two different approaches: we try early data fusion using one SOM, then we use supervised perceptrons to learn the multimodal representations based on the two unimodal SOMs activities.

\subsubsection{SOMs vs. SNNs approaches for unsupervised learning}
\label{sec_som-vs-snn}

\begin{table*}[ht]
\centering
\caption{Digits classification comparison}
\label{tab_digits_accuracy}
\begin{center}
\resizebox{0.9\linewidth}{!}{
\begin{tabular}{l l c c l l c}
\hline
\textbf{ANN} & \textbf{Model}           & \textbf{Neurons} & \textbf{Labels (\%)} * & \textbf{Modality}   & \textbf{Dataset}        & \textbf{Accuracy (\%)}  \\ \hline
\multirow{4}{*}{SNN}  & Diehl et al. \cite{diehl2015stdp} (2015)  & 400        & 100                 & Unimodal   & MNIST          & 88.74          \\
                      & Hazan et al. \cite{hazan2018lmsnn} (2018)  & 400        & 100                 & Unimodal   & MNIST          & 92.56          \\
                      & Rathi et al. \cite{rathi2018multimodal_stdp} (2018) & 400        & 100                 & Unimodal   & MNIST          & 86.00          \\ 
                      & Rathi et al. \cite{rathi2018multimodal_stdp} (2018)  & 400        & 100                 & Multimodal & \makecell{MNIST + \\ TI46}   & 89.00          \\ \hline
SOM                   & Khacef et al. [this work] (2020) & 356        & 1                   & Multimodal & \makecell{MNIST + \\ SMNIST} & \textbf{95.07} \\ \hline
\end{tabular}
}
\end{center}
\begin{flushleft}
    \footnotesize{$^*$ Labeled data are only used for the neurons labeling after unsupervised training.}
\end{flushleft}
\end{table*}

Table \ref{tab_digits_accuracy} summarizes the digits classification accuracy achieved using brain-inspired unsupervised approaches, namely SOMs with self-organization (Hebb, Oja and Kohonen principles) and SNNs with STDP. We achieve the best accuracy with a gain of about $6 \%$ over Rathi and Roy \cite{rathi2018multimodal_stdp}, which is to the best of our knowledge the only work that explores brain-inspired multimodal learning for written/spoken digits classification. It is to note that we do not use the TI46 spoken digits database \cite{liberman1991ti46} (not freely available), but a subpart of Google Speech Google Speech Commands \cite{warden2018speech_commands} as presented in section \ref{sec_wr-sp-digits}.
We notice that all other works use the complete training dataset to label the neurons, which is incoherent with the goal of not using labels, as explained in \cite{khacef2019self-organizing_neurons}. Moreover, the work of Rathi and Roy \cite{rathi2018multimodal_stdp} differs from our work in the following points:

\begin{itemize}
    \item The cross-modal connections are formed randomly and initialized with random weights. The multimodal STDP learning is therefore limited to connections that have been randomly decided, which induces an important variation in the network performance.
    
    \item The cross-modal connections are not bi-directional, thus breaking with the biological foundations of reentry and CDZ. Half the connections carry spikes from image to audio neurons and the other half carry spikes from audio to image neurons, otherwise making the system unstable.
    
    \item The accuracy goes down beyond $26\%$ connections. When the number of random cross-modal connections is increased, the neurons that have learned different label gets connected. We do not observe such a behavior in the ReSOM, as shown in Figure \ref{fig_cdz-prune}.
    
    \item The SNN computation is distributed, but requires an all-to-all connectivity amongst neurons. This full connectivity goes against the scalability of the network as discussed in section \ref{sec_iterative-grid}.
    
    \item The decision of the multimodal network is computed by \textit{observing} the spiking activity in both ensembles, thus requiring a central unit.
\end{itemize}

Nevertheless, the STDP-based multimodal learning is still a promising approach for the hardware efficiency of SNNs \cite{khacef2018mlp_vs_snn}, and because of the alternative they offer for using even-based sensors with asynchronous computation.

\subsubsection{SOM early data fusion}
We find in the literature two main different strategies for multimodal fusion \cite{baltrusaitis2019multimodal_ml} \cite{cholet2019associative_memory}: (1) score-level fusion where data modalities are learned by distinct models then their predictions are fused with another model that provides a final decision, and (2) data-level fusion where modalities are concatenated then learned by a unique model. 
Our approach can be classified as a classifier-level fusion which is closer to score-level fusion and usually produces better results than feature-level or data-level fusion for classification tasks \cite{guo2014fusion-acc-phy} \cite{peng2016hierarchical} \cite{biagetti2018fusion-acc-semg}. However, it is worth trying to learn the concatenated modalities with one SOM having as much neurons as the two uni-modal SOMs, for a fair comparison.
We use $361$ and $529$ neurons for digits and hand gestures respectively. We have few neurons more compared to the sum of the two uni-modal SOMs, as we want to keep the same square grid topology. We train the SOMs with the same hyper-parameters as for the uni-modal SOMs, and reach $90.68\% \pm 0.29$ and $75.6\% \pm 0.32$ accuracy for digits and hand gestures, respectively. We still have a gain compared to the uni-modal SOMs, but we have an important loss of $-4.39\%$ for digits and a negligible loss of $-0.13\%$ for hand gestures compared to the proposed ReSOM multimodal association. 
The incremental aspect of the ReSOM from simple (unimodal) to more complex (multimodal) representations improves the system's accuracy, which is coherent with the literature findings.
Furthermore, the accuracy is not the only metric, as the memory footprint is an important factor to take in consideration when choosing a fusion strategy \cite{castanedo2013review_fusion}, especially for embedded systems. Indeed, since we target a hardware implementation on FPGA, the total number of afferent and lateral synaptic weights are parameters that require on-chip memory, which is very limited. With a simple calculation using the number of neurons and input dimensions, we find that we have a gain of $49.84\%$ and $40.96\%$ for digits and hand gestures respectively using the multimodal association compared to a data-level fusion strategy.

\subsubsection{SOMs coupled to supervised fusion}
In order to have an approximation of the best accuracy that we could obtain with multimodal association, we used a number of perceptrons equal to the number of classes on top of the two uni-modal SOMs of the two databases, and performed supervised learning for the same number of epochs ($10$) using gradient descent (Adadelta algorithm). 
We obtain $91.29\% \pm 0.82$ and $80.19\% \pm 0.63$ of accuracy for the digits and hand gestures respectively. Surprisingly, we have a loss of $-3.78\%$ for the digits. However, we have a gain of $4.43\%$ for the hand gestures.
We argue that the hand gestures dataset is too small to construct robust multimodal representations through unsupervised learning, and that could explain the smaller overall gain compared to the digits dataset.


\section{Discussion} \label{sec_discussion}

\subsection{A universal multimodal association model?}
The development of associations between co-occurring stimuli for multimodal binding has been strongly supported by neurophysiological evidence \cite{fiebelkorn2010dual_mechanism} \cite{ursino2014neurocomputational_review}. Similar to \cite{vavrecka2013multimodal_connecionist}, \cite{morse2015infants_map} and \cite{parisi2017multimodal_action} and based on our experimental results, we argue that the co-occurrence of sensory inputs is a sufficient source of information to create robust multimodal representations with the use of associative links between unimodal representations that can be incrementally learned in an unsupervised fashion.

In terms of learning, the best methods are based on $Hebb$'s learning with a slightly better accuracy over $Oja$'s learning, but the overall results are more homogeneous using $Oja$'s learning that prevents the synaptic weights from growing indefinitely. The best results are obtained using $Hebb-Max_{Norm}^{BMU}$ with $95.07\% \pm 0.08$ and $Hebb-Sum_{Norm}^{All}$ with $75.73\% \pm 0.91$ for the digits and hand gestures datatabases, respectively. We notice that the $BMU$ method is coupled with the $Max$ update while the $All$ neurons method is coupled with the $Sum$ update, and the $Norm$ activities usually perform better than $Raw$ activities. However, we cannot have a final conclusion on the best method, especially since it depends on the nature of the dataset.

Moreover, the experimental results depend on the $\beta$ hyper-parameter in Equation \ref{eq_cdz_gaussian}, the Gaussian kernel width that has to be tuned for every database and every method. Thanks to the multiplicative update, the values of both SOMs are brought into the same scale which gives the possibility to elect the correct global BMU, and we get rid of a second hyper-parameter that would arise with a sum update method like in \cite{jayaratne2018multi-sensory_fusion}. However, it is still time-taking in the exploration of the proposed methods for future works, even if it is a common limit when dealing with any ANN. Finding a more efficient approach for computing $\beta$ is part of our ongoing works.

Finally, multimodal association bridges the gap between unsupervised and supervised learning, as we obtain approximately the same results compared to MNIST using a supervised Multi-Layer Perceptron (MLP) with $95.73 \%$ \cite{khacef2018mlp_vs_snn} and S-MNIST using a supervised attention Recurent Neural Network (RNN) with $94.5 \%$ \cite{andrade2018arnn_speech} (even though this results was obtained on 20 commands). Multimodal association can also be seen as way to reach the same accuracy of about $95\%$ as \cite{diehl2015stdp} with much less neurons, going from $6400$ neurons to $356$ neurons, i.e. a gain of $94\%$ in the total number of neurons.
It is therefore a very promising approach to deeper explore, as we have in most cases the possibility to include multiple sensory modalities when dealing with the real-world environment.

\subsection{SOMA: Toward hardware plasticity}
This work is part of the Self-Organizing Machine Architecture (SOMA) project \cite{khacef2018neuromorphic_hardware}, where the objective is to study neural-based self-organization in computing systems and to prove the feasibility of a self-organizing multi-FPGA hardware structure based on the IG cellular neuromorphic architectures.
In fact, the concept of the IG is supported in \cite{heylighen2003self-organization} as it states that \say{changes initially are local: components only interact with their immediate \textit{neighbors}. They are virtually independent of components farther away. But self-organization is often defined as global order emerging from local interactions}. Moreover, it states that \say{a self-organizing system not only regulates or adapts its behavior, it creates its own organization. In that respect it differs fundamentally from our present systems, which are created by their designer}. 

Indeed, the multimodal association through Hebbian-like learning is a self-organization that defines the inter-SOMs structure, where neurons are only connected to each other when there is a strong correlation between them. That's a form of \textit{hardware plasticity}.
The hardware gain of the ReSOM self-organization is therefore the gain in communication support, which is proportional to the percentage of remaining synapses for each neuron after learning and pruning that reduces the number of connections, thus the number of communications and therefore the overall energy consumption.
Hence, the system is more energy-efficient as only relevant communications are performed without any control by an external expert.


\section{Conclusion and further works} 
\label{sec_conclusion}
We proposed in this work a new brain-inspired computational model for multimodal unsupervised learning called the ReSOM.
Based on the reentry paradigm proposed by Edelman, it is a generic model regardless of the number of maps and the number of neurons per map. The ReSOM learns unimodal representations with Kohonen-based SOMs, then creates and reinforces the multimodal association via sprouting, Hebbian-like learning and pruning. It enables both structural and synaptic plasticities that are the core of neural self-organization. We exploited both convergence and divergence that are highlighted by Damasio thanks to the bi-directional property of the multimodal representation in a classification task: the divergence mechanism is used to label one modality based on the other, and the convergence is used to introduce cooperation and competition between the modalities and reach a better accuracy than the best of the two unimodal accuracies. 
Indeed, our experiments show that the divergence labeling leads to approximately the same unimodal accuracy as when using labels, and we reach a gain in the multimodal accuracy of $+8.03\%$ for the written/spoken digits database and $+5.67\%$ for the DVS/EMG hand gestures database. Our model exploits the natural complementarity between different modalities like sight and sound as shown by the confusion matrices, so that they complete each other and improve the multimodal classes separability.
Implemented on the IG cellular neuromorphic architecture, the ReSOM's inter-map structure is learned along the system's experience through self-organization and not fixed by the user. It leads to a gain in the communication time which is proportional to the number of pruned lateral synapses for each neuron, which is about $80\%$ of the possible connections. 
In addition to the convergence and divergence gains, the ReSOM self-organization induces a form of hardware plasticity which has an impact on the hardware efficiency of the system, and that's a first result that opens very interesting perspectives for future designs and implementations of self-organizing architectures inspired from the brain's plasticity.


\section*{Conflict of Interest Statement}
The authors declare that the research was conducted in the absence of any commercial or financial relationships that could be construed as a potential conflict of interest.

\section*{Author Contributions}
LK, LR and BM conceived the idea. LK developed the code and performed the experiments, LR and BM supervised the work. LK, LR and BM wrote the paper.

\section*{Funding}
This material is based upon work supported by the French Research Agency (ANR) and the Swiss National Science Foundation (SNSF) through SOMA project ANR-17-CE24-0036.

\section*{Acknowledgments}
This manuscript has been released as a pre-print at https://arxiv.org/abs/2004.05488 \cite{khacef2020cdz}.
The authors would like to acknowledge the 2019 Capocaccia Neuromorphic Workshop and all its participants for the fruitful discussions.

\section*{Data Availability Statement}
The datasets for this study can be found in \cite{khacef2019multimodal_digits} and \cite{ceolini2019hand_gestures}.

\bibliographystyle{IEEEtran}
\bibliography{biblio}

\end{document}